\documentclass[a4paper,conference]{IEEEtran}
\IEEEoverridecommandlockouts
\ifCLASSINFOpdf
\else
\fi
\usepackage{graphicx}
\usepackage{comment}
\usepackage{amsmath,amssymb} 
\usepackage{color}
\usepackage{wrapfig}
\usepackage{multirow}
\usepackage{colortbl}
\usepackage{epsfig}
\usepackage{graphicx}
\usepackage{subfigure}
\usepackage{xcolor}

\newcommand{\yq}[1]{[{\color[rgb]{1.00,0.00,0.50} #1}]}

\newcommand{\Eq}[1]{Eq.~(\ref{eq:#1})}
\newcommand{\eq}[1]{\Eq{#1}}

\newcommand{\fig}[1]{Fig.~\ref{fig:#1}}
\newcommand{\tab}[1]{Table~\ref{tab:#1}}
\newcommand{\tabl}[1]{Table~\ref{table:#1}}

\usepackage{graphicx}
\usepackage{amsmath}
\usepackage{amssymb}
\usepackage{booktabs}

\usepackage{xspace}

\usepackage[marginal]{footmisc}
\usepackage[pagebackref=true,breaklinks=true,letterpaper=true,colorlinks,bookmarks=false]{hyperref}
\usepackage[T1]{fontenc}

\begin{document}
%
\title{WeightAlign: Normalizing Activations by \\ Weight Alignment}

\author{\IEEEauthorblockN{Xiangwei Shi$^\star$, Yunqiang Li$^\star$, Xin Liu$^\star$ and Jan van Gemert\thanks{$^\star$ The authors contributed equally to this work.}} 
\IEEEauthorblockA{Computer Vision Lab\\
Delft University of Technology, The Netherlands
}

}


%


\maketitle

\begin{abstract}
Batch normalization (BN) allows training very deep networks by normalizing activations by mini-batch sample statistics which renders BN unstable for small batch sizes. Current small-batch solutions such as Instance Norm, Layer Norm, and Group Norm use channel statistics which can be computed even for a single sample. Such methods are less stable than BN as they critically depend on the statistics of a single input sample. To address this problem, we propose a normalization of activation without sample statistics. We present WeightAlign: a method that normalizes the weights by the mean and scaled standard derivation computed within a filter, which normalizes activations without computing any sample statistics. Our proposed method is independent of batch size and stable over a wide range of batch sizes. Because weight statistics are orthogonal to sample statistics, we can directly combine WeightAlign with any method for activation normalization. We experimentally demonstrate these benefits for classification on CIFAR-10,  CIFAR-100, ImageNet, for semantic segmentation on PASCAL VOC 2012 and for domain adaptation on Office-31.
\end{abstract}


%
\IEEEpeerreviewmaketitle

\section{Introduction}

Batch Normalization~\cite{ioffe2015batch} is widely used in deep learning. Examples include image classification~\cite{chan2015pcanet,he2015delving,krizhevsky2012imagenet}, object detection~\cite{he2017mask,redmon2016you,ren2015faster}, semantic segmentation~\cite{long2015fully,noh2015learning,ronneberger2015u}, generative models~\cite{arjovsky2017wasserstein,goodfellow2014generative,radford2015unsupervised}, \textit{etc}. It is fair to say that for optimizing deep networks, BatchNorm is truly the norm.

BatchNorm stabilizes network optimization by normalizing the activations during training and exploits mini-batch sample statistics. The performance of the normalization thus depends critically on the quality of these sample statistics. Having accurate sample statistics, however, is not possible in all applications. An example is domain adaptation, where the statistics of the training domain samples do match the target domain statistics, and alternatives to BatchNorm are used.~\cite{chang2019domain,li2016revisiting}.
High resolution images are another example, as used in object detection~\cite{girshick2015fast,he2017mask,ren2015faster},
segmentation~\cite{long2015fully,noh2015learning} and video recognition~\cite{carreira2017quo,tran2015learning}, where only a few or just one sample per mini-batch fits in memory. For these cases, it is difficult to accurately estimate activation normalization statistics from the training samples.

To overcome the problem of unreliable sample statistics, various normalization techniques have been proposed which make use of other statistics derived from samples, such as layer~\cite{ba2016layer}, instance~\cite{ulyanov2016instance} or group~\cite{wu2018group}. These methods are applied independently per sample and achieve good performance, but gather statistics just on a single sample and are thus less reliable than BatchNorm.

\begin{figure*}[t]
	\centering
	\includegraphics[width= 0.92\textwidth]{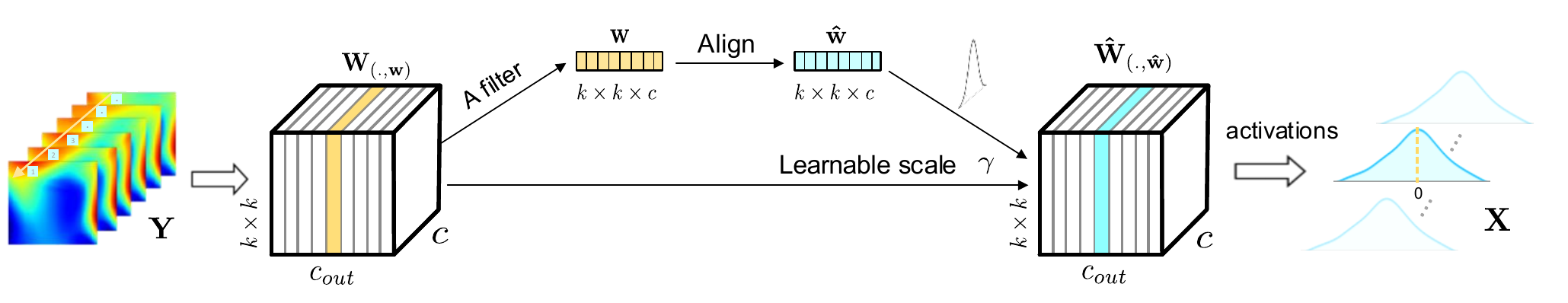}
	\caption{Overview of WeightAlign (WA): Aligning filter weights allows normalizing channel activations. 
	The weight $\textbf{W}$ of an arbitrary convolutional layer is composed with $C_{out}$  filters with size $C\times k\times k$.
Within a filter (yellow region in the left side), we first flatten it to a vector and then apply our proposed WA in \eq{reparmeterize1} to normalize the filter weights with a introduced learnable parameter $\gamma$.
Normalizing weights of each filter  in forward pass can realize the normalization of activations within each channel. Details in Section \ref{approach}. 
	}
	\label{fig:distribution}
\end{figure*}

In this paper, we propose WeightAlign: normalizing activations without using sample statistics. Instead of sample statistics, we re-parameterize the weights within a filter to arrive at correctly normalized activations.  See~\fig{distribution} for a visual overview. Our method is based on weight statistics and is thus orthogonal to sample statistics. This allows us to exploit two orthogonal sources
to normalize activations: 
the traditional one based on sample statistics in combination with our proposed new one based on weight statistics. We have the following contributions.
\begin{itemize}
	\setlength\itemsep{0mm}
	\item WeightAlign: A new method to normalize filter weights.
	\item Activation normalization without computing sample statistics.
	\item Performance independent of batch size, and stable performance over a wide range of batch sizes.
	\item State-of-the-art performance on 5 datasets in image classification, object detection, segmentation and domain adaptation. 
\end{itemize}


\section{Related Work}
 \noindent \textbf{Network Normalization by Sample Statistics. }
The idea of whitening the input to improve training speed~\cite{lecun2012efficient}
can be extended beyond the input layer to intermediate representations inside the network. Local Response Normalization~\cite{krizhevsky2012imagenet}, used in AlexNet, normalizes the activations in a small neighborhood for each pixel. Batch Normalization (BN)~\cite{ioffe2015batch} performs normalization using sample statistics computed over mini-batch, which is helpful for training very deep networks. BN unfortunately suffers from performance degradation when the statistical estimates become unstable for small batch-size based tasks.

To alleviate the small batches issue in BN, Batch Renormalization~\cite{ioffe2017batch} introduces two extra parameters to correct the statistics during training. However, Batch Renormalization is still dependent on mini-batch statistics, and degrades performance for smaller mini-batches. EvalNorm~\cite{singh2019evalnorm} corrects the batch statistics during inference procedure. But it fails to fully alleviate the small batch issue. SyncBN~\cite{peng2018megdet} handles the small batch problem by computing the mean and variance across multiple GPUs which is not possible without having multiple GPUs available.

Because small mini-batch sample statistics are unreliable, several methods~\cite{ba2016layer,ulyanov2016instance,wu2018group} perform feature normalization based on channel sample statistics. Instance Normalization~\cite{ulyanov2016instance} performs normalization similar to BN but only for a single sample. Layer Normalization~\cite{ba2016layer} uses activation normalization along the channel dimension for each sample. Filter Response Normalization~\cite{singh2020filter}
proposes a novel combination of a normalization that operates on each activation channel of each batch element independently. Group Normalization~\cite{wu2018group} divides channels into groups and computes the mean and variance within each group for normalization.
Local Context Normalization~\cite{ortiz2020local} normalizes every feature based on the filters in its group and a window around it. 
These methods can alleviate the small batch problem to some extent, yet have to compute statistics for just a single sample both during training and inference which introduces instabilities~\cite{shao2019ssn,yan2020towards}.
Instead, in this paper we cast the activation normalization over \emph{feature space}
into weights manipulation over \emph{parameter space}. Network parameters are independent of any sample statistics and makes our method particularly well suited as an orthogonal information source which can compensate for unstable sample normalization methods.

 \noindent \textbf{Importance of weights.}
Proper network weight parameter initialization~\cite{he2016deep,krizhevsky2012imagenet,simonyan2014very} is essential for avoiding vanishing and exploding gradients~\cite{pascanu2013difficulty}. Randomly initializing weight values from a normal distribution~\cite{krizhevsky2012imagenet} is not ideal because of the stacking of non-linear activation functions. Properly scaling the initialization for stacked sigmoid or tanh activation functions leads to Xavier initialization~\cite{glorot2010understanding}, but it is not valid for ReLU activations. He \textit{et al.}~\cite{he2015delving} extend Xavier initialization~\cite{glorot2010understanding} and derive a sound initialization for ReLU activations.
For hypernets, \cite{chang2019principled} developed principled techniques for weight initialization to solve the exploding
activations even with the use of BN.
In \cite{mishkin2015all} a data-dependent initialization is proposed to mimic the effect of BN  to normalize the variance per layer to one in the first forward pass.
 To address the initialization problem of residual nets,
Zhang \textit{et al.}~\cite{zhang2019fixup} propose fixup initialization.
We draw inspiration from these weight initialization methods to derive our weight alignment for activation normalization.

Weight decay~\cite{krogh1992simple} adds a regularization term to encourage small weights. Max-norm regularization is used in \cite{srebro2005rank} to avoid extremely large weights. Instead of introducing an extra term to the loss, Weight Norm~\cite{salimans2016weight}  re-parameterizes weights by dividing its norm to accelerate convergence. Such a re-parameterization, however, may greatly magnify the magnitudes of input signals. In contrast, we scale each filter's weights as  inspired by He \textit{et al.}~\cite{he2015delving}'s weight initialization approach, to generate activations with zero mean and maintained variance. This allows us to normalize activations without using any sample statistics.

\section{Proposed Method} \label{approach}
Batch Normalization (BN)~\cite{ioffe2015batch} normalizes the features in a single channel to a zero mean and unit variance distribution, as in \fig{pSensitivity}, to stabilize the optimization of deep networks. The normalized features  $\mathbf{\hat{x}}$ are computed channel-wise at training time using the sample mean $\mu_\beta$ and standard derivation $\sigma_\beta$ over the input features $\mathbf{{x}}$ as:
 \begin{equation}
\mathbf{\hat{x}} = \frac{\mathbf{{x}} - \mu_\beta}{\sigma_\beta},
\label{eq:BN_eq}
\end{equation}
where $\mu_\beta$ and $\sigma_\beta$ are functions of \emph{sample statistics} of $\mathbf{{x}}$.
For each activation $\mathbf{\hat{x}}$, a pair of trainable parameters $\gamma$, $\beta$ are introduced to scale and shift the normalized value that 
\begin{equation}
    \mathbf{r} =\gamma \mathbf{\hat{x}} + \beta.
    \label{eq:gamma}
\end{equation}
 BN uses the  running averages of sample statistics within each mini-batch to reflect the statistics over the full training set. A small batch leads to inaccurate estimation of the sample statistics, thus using small batches degrades the accuracy severely.

In the following, we express the sample statistics in terms of filter weights to eliminate the effect of small batch, and re-parameterize the weights within each filter to realize the normalization of activations within each channel.

\subsection{Expressing activation statistics via weights}

For a convolution layer, let $\mathbf{x}$ be a single channel output activation, $\mathbf{Y}$ be the input activations, and $\mathbf{w}$, $b$ be the corresponding filter and bias scalar.
Specifically, a filter $\mathbf{w}$  consists of $c$ channels with a spatial size of $k\times k$.
To normalize activations by using weights, we show how the sample statistics $\mu_\beta$ and $\sigma_\beta$ in \eq{BN_eq} are expressed in terms of filter weights.

An individual response value ${x}(t)$ in $\mathbf{x}$ is a sum of the product between  the values in $\mathbf{w}$ and
the values in a co-located $k^2c$-by-$1$ vector $\mathbf{y}(t)$ in $\mathbf{Y}$, which is equivalently  written as the dot product of filter $\mathbf{w}$ and vector $\mathbf{y}(t)$,
 \begin{equation}
x(t) = \mathbf{w} \cdot \mathbf{y}(t) + b.
\label{eq:aresponse}
\end{equation}
By computing each individual response $x(t)$ at any location $t$, we obtain the full single channel output activation $\mathbf{x}$.

We use random variables $x$, $Y$ and $w$ to present each of the elements in $\mathbf{x}$,  $\mathbf{Y}$ and $\mathbf{w}$  respectively. Following~\cite{chang2019principled,glorot2010understanding,he2015delving}, we make the following assumptions about the network:  (\textbf{1}) The $w$ , ${Y}$ and $b$
are all independent of each other. (\textbf{2}) The bias term $b$ equals 0.

\textit{Expressing $\mu_\beta$ of \eq{BN_eq} in terms of weights:} The mean  of activation $\mathbf{x}$ can be equivalently represented via filter weights as:
\begin{equation}
{\mu_\beta =\mathbb{E}}[x] = n \mathbb{E}[wY] = n \mathbb{E}[w] \mathbb{E}[Y],
\label{eq:1mean}
\end{equation}
where $\mathbb{E}$ is the expected value and $n = k^2c$ denotes the number of weight values in a filter. 

\textit{Expressing $\sigma_\beta$ of \eq{BN_eq} in terms of weights:} The variance of activation $\mathbf{x}$ is:
\begin{gather}
\begin{aligned}
\sigma_\beta^2 = \mathrm{Var}[x]  = & n {\mathrm{Var}}[wY] = n({\mathbb{E}}[w^2 Y^2] - \mathbb{E}^2[w Y]) \\
 =
& n({\mathbb{E}}[w^2] {\mathbb{E}}[Y^2] - \mathbb{E}^2[w] \mathbb{E}^2[Y])
\end{aligned}
\label{eq:1var}
\end{gather}

With the \eq{1mean} and \eq{1var}, we will manipulate the weights to normalize activations in the following sections.

\subsection{WeightAlign}
\label{weightalignEq}
We normalize activations by encouraging the mean $\mu_\beta$ of the activations in \eq{1mean} to be zero, and maintaining the variance of the activations to be the same in all layers. Inspired by~\cite{he2015delving}, we manipulate the weights to satisfy these requirements.


The mean of the activations $\mu_\beta$ in \eq{1mean} is forced to be zero when the weight $w$ in a filter has zero mean:
\begin{equation}
\mathbb{E}[w]  = 0.
\label{eq:2mean}
\end{equation}

We can exploit $\mathbb{E}[w]  = 0$ in \eq{1var} to further simplify the variance of the activations $\sigma_\beta^2$ in terms of the variance of the weights as,
 \begin{equation}
\sigma_\beta^2=\mathrm{Var}[x] = n{\mathbb{E}}[w^2] {\mathbb{E}}[Y^2] = n {\mathrm{Var}}[w]\mathbb{E}[Y^2].
\label{eq:2var}
\end{equation}
Thus, the activation variance $\sigma_\beta^2$ depends on the variance of the weights $w$ and on the current layer input $Y$. To simplify further, we follow~\cite{he2015delving} and assume a ReLU activations function. We use $Z$ to define the outputs of the previous layer before the activation function, where $Y = \mathrm{ReLU}(Z)$. 

If the weight of the previous layer has a symmetric distribution around zero 
then $Z$ 
has a symmetric distribution around zero.\footnote{See proof in the supplementary material.} Because the ReLU sets all negative values to 0, the variance is halved~\cite{he2015delving}, and we have that $\mathbb{E}[Y^2] =\frac{1}{2} \mathrm{Var}[Z]$. Substituted into \eq{2var}, we have
 \begin{equation}
\sigma_\beta^2=\mathrm{Var[x] = \frac{1}{2} n {\mathrm{Var}}[w]{\mathrm{Var}}[Z]}.
\label{eq:3var}
\end{equation}
This shows the relationship of variance between the previous layer's output before the non-linearity $Z$ and the current layer's single channel activation $x$. 

Our goal is to manipulate the weights to keep the activation variance of all layers normalized. Thus we relate the variance of this layer $\sigma_\beta^2$ with the variance in the previous layer $\mathrm{Var[Z]}$, which can be achieved in terms of the weights by setting $\frac{1}{2} n \mathrm{Var}[w]$ in \eq{3var} to  a proper scalar (\emph{e.g.} 1):
\begin{equation}
\frac{1}{2} n \mathrm{Var}[w] = 1.
\label{eq:con0}
\end{equation}

With \eq{2mean} and \eq{con0}, we can re-parameterize a single filter weights to have zero mean and a standard deviation $\sqrt{{2}/{n}}$, which achieves the activation normalization like BN in \eq{BN_eq}. Thus, here we propose WeightAlign as,
 \begin{equation}
\hat{{w}} = \gamma \ \frac{{w}- \mathbb{E}[{w}]}{\sqrt{{n}/{2} \cdot \mathrm{Var}[{w}]}  },
\label{eq:reparmeterize1}
\end{equation}
where  $\gamma$ is a learnable scalar parameter 
similar to the scale parameter in BN~\cite{ioffe2015batch} like \eq{gamma}. Since WA manipulates the weights, it does not have the $\beta$ term as in the form of BN, which is applied on activations directly.
The detail explanation about this is given in supplementary material.
From \eq{reparmeterize1}, we can tell that the proposed WeightAlign(WA) method does not rely on sample statistics computed over a mini-batch, which makes it independent of batch size.

\subsection{Initialization of weights}
We initialize each filter weights  $\mathbf{w}$ to be a zero-mean symmetric Gaussian distribution whose standard deviation (std) is $\sqrt{{2}/{n}}$ where $n = k^2 c$. For the first layer, we should have $\sqrt{{1}/{n}}$ since the ReLU activation is not applied on the input. But the factor ${1}/{2}$ can be neglected if it just exists on one layer and for simplicity, we use the same deviation  $\sqrt{{2}/{n}}$ for initialization.

\begin{figure*}[t!]
 \centering
\begin{tabular}{rccccc}
\rotatebox{90}{\ \ \ \ \ \ \scriptsize{3rd layer}} & \includegraphics[width=0.17\textwidth]{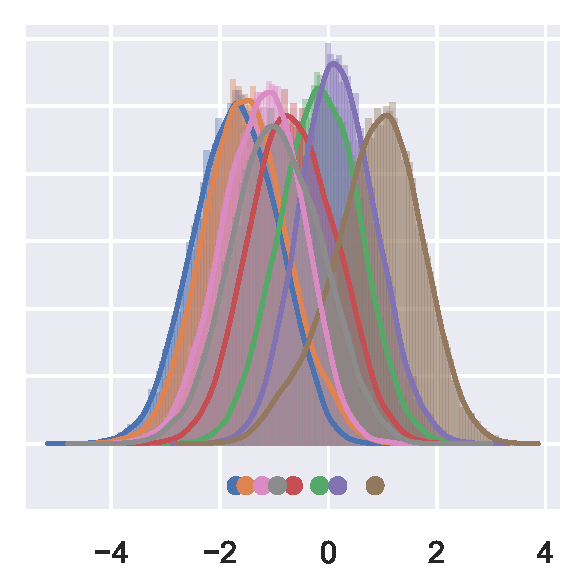} &
\includegraphics[width=0.17\textwidth]{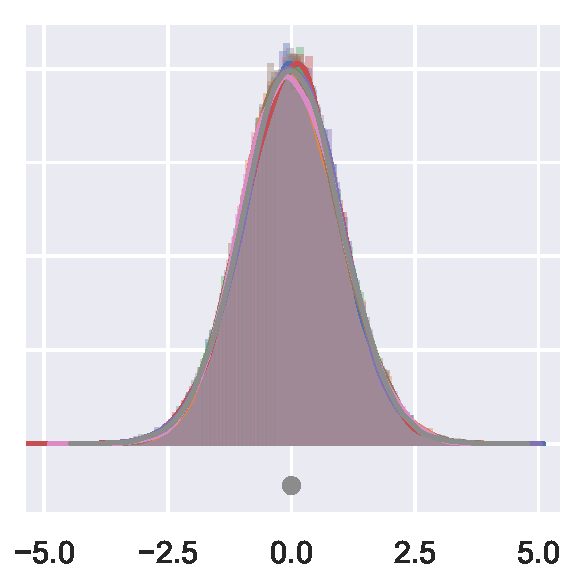} &
\includegraphics[width=0.17\textwidth]{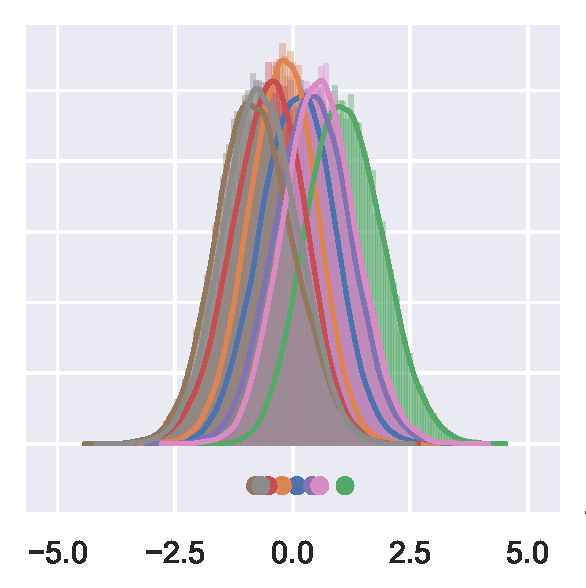} &
\includegraphics[width=0.17\textwidth]{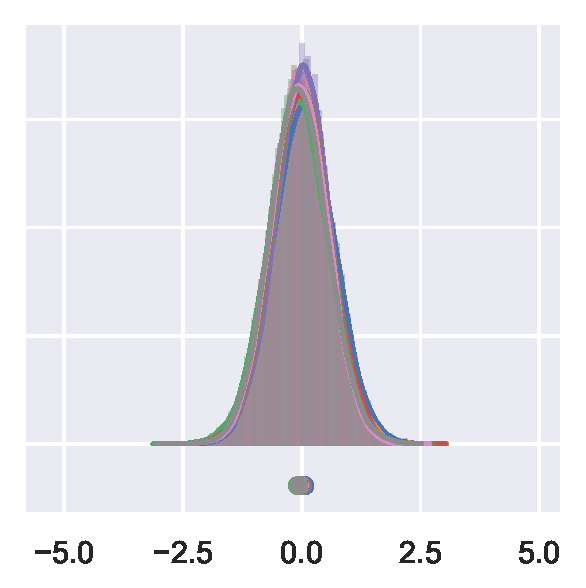}&
\includegraphics[width=0.17\textwidth]{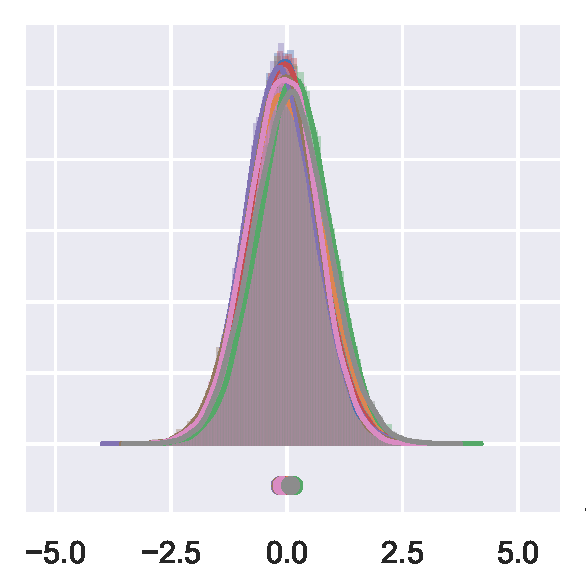} \\
\rotatebox{90}{\ \ \ \ \scriptsize{classifier layer}} &
\includegraphics[width=0.17\textwidth]{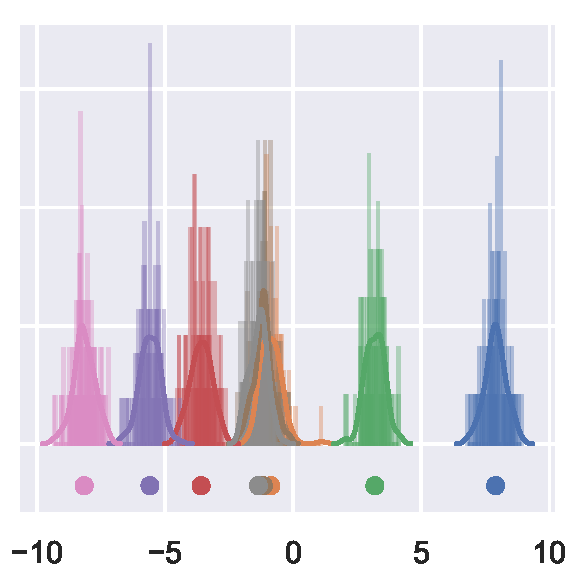} &
\includegraphics[width=0.17\textwidth]{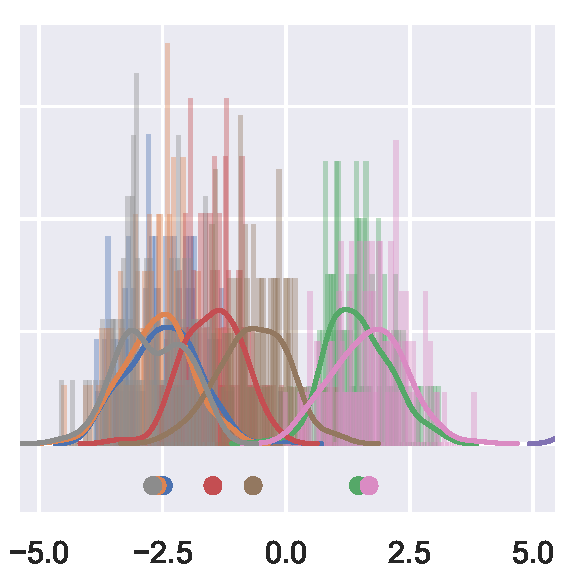} &
\includegraphics[width=0.17\textwidth]{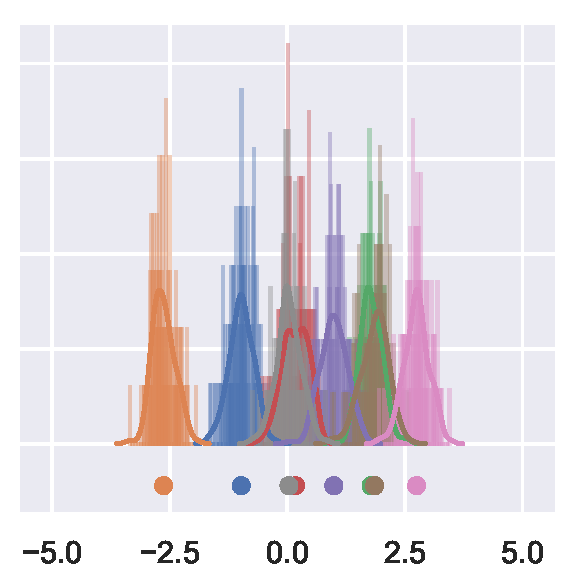} &
\includegraphics[width=0.17\textwidth]{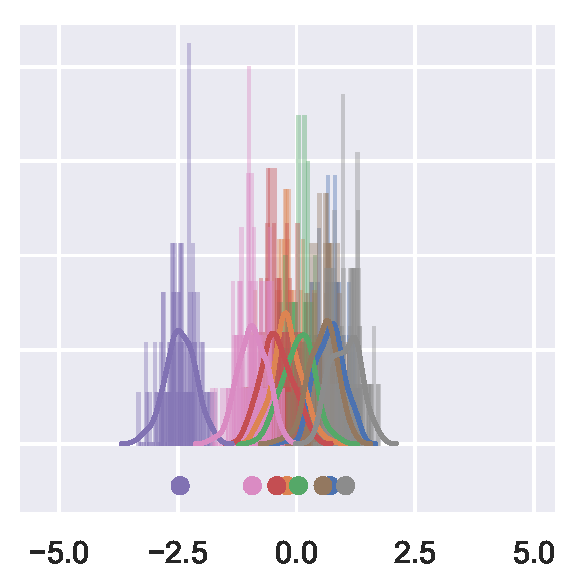}&
\includegraphics[width=0.17\textwidth]{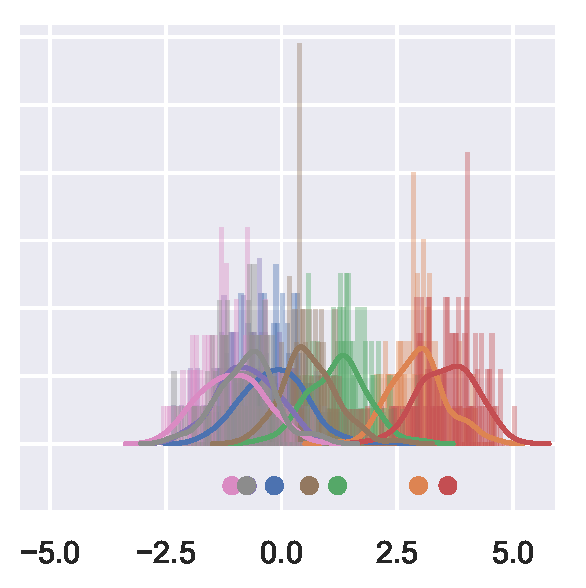} \\
& \scriptsize{(a) Baseline} &   \scriptsize{(b) BN} & \scriptsize{(c) GN }& \scriptsize{(d) WA} &  \scriptsize{(e) WA+GN} \\
\end{tabular}
\caption{ 
Our WA method can alleviate internal covariate shift of activation (see Section~\ref{Analysis} for details).
Each color represents the activation distribution of different channels for different layers. 
For baseline model, the 'Blue' indexed channel will dominate all other channels, leading to a constant classification result. Normalizing channel activation in intermediate layer can avoid the constant output for the final classifier layer, which can be realized by WA.
}
 \label{fig:pSensitivity}
\end{figure*}

\subsection{Empirical analysis and examples} \label{Analysis}




To show the distributions of channel activations using different methods, we build an 8-layer deep network containing 7 convolutional layers and one classification layer with ReLU as nonlinear activation function. A mini-batch of size 128 independent data is sampled from $\mathcal{N}(0, 1)$ is used as input.
Specifically, the standard Kaiming initialization~\cite{he2015delving} without bias term is used to initialize the weights. \fig{pSensitivity} shows the activation distributions of 8 different channels taken from two layers, the 3rd intermediate convolutional layer and the last classifier layer before the softmax loss. Each channel of the last classifier layer represents one class.
To exclude the training effect, we first use the initialization model to conduct comparison between sample statistics based normalization and our WA normalization.  
For the visualization of trained model, please see Section \ref{experiment}. 

For the baseline model, it can be seen in \fig{pSensitivity}(a) that the activation distributions in the intermediate layer start drifting, leading  to  a  constant  output  for  the  classifier  layer (\emph{i.e.} the 'Blue' indexed class/channel).
From \fig{pSensitivity}(b) of Batch Norm (BN)~\cite{ioffe2015batch}, we note that reducing the Internal Covariate Shift (ICS) in intermediate layer can alleviate the distribution drifting for the final classifier layer. The output of the classifier layer is no longer a constant function. Group Norm(GN)~\cite{wu2018group} can also reduce ICS to some extent, as shown in \fig{pSensitivity}(c).

As shown in \fig{pSensitivity}(d), WA can realize similar functionality of BN.
Our proposed WeighAlign (WA) re-parameterizes the weights within each filter in~\eq{reparmeterize1} to normalize channel activation.
Since weight statistics is orthogonal to sample statistics, WA can be used in conjunction with BN, GN, LN and IN, as shown in \fig{pSensitivity}(e).
Please refer to supplementary material for visualization of other normalization methods.

\section{Experiments} \label{experiment}
We show extensive experiments on three tasks and five datasets:
image classification on CIFAR-10, CIFAR-100~\cite{krizhevsky2009learning} and
ImageNet; domain adaptation on Office-31; and semantic segmentation on PASCAL VOC 2012~\cite{everingham2007pascal} where we evaluate VGG~\cite{simonyan2014very} and residual networks as ResNet~\cite{he2016deep}.

\subsection{Datasets}
\noindent \textbf{CIFAR-10 \& CIFAR-100.} {CIFAR-10} and {CIFAR-100}~\cite{krizhevsky2009learning} consist of 60,000 32$\times$32 color images with 10 and 100 classes, respectively. For both datasets, 10,000 images are selected as the test set and the remaining  50,000  images are used for training. We perform image classification experiments and ablation study on these datasets.

 \noindent \textbf{ImageNet.} ImageNet~\cite{deng2009imagenet} is a classification dataset with 1,000 classes. The size of the training dataset is around 1.28 million, and 50,000 validation images are used for evaluation.

\noindent \textbf{Office-31.} Office-31~\cite{Office-31} is a dataset for domain adaptation with 4,652 images with 31 categories. The images are collected from three distinct domains: Amazon (A), DSLR (D), and Webcam (W). The largest domain, Amazon, has 2,817 labeled images. The 31 classes consists of objects commonly encountered in office settings, such as file cabinets, keyboards and laptops.

\noindent \textbf{PASCAL VOC 2012.} PASCAL VOC 2012~\cite{everingham2007pascal} contains a semantic segmentation set and includes 20 foreground classes and a single background class. The original segmentation dataset contains 1,464 images for training, and 1,499 images for evaluation. We use the augmented training dataset including 10,582 images~\cite{hariharan2011semantic}.

\subsection{Implementation details}
We use Stochastic Gradient Descent(SGD) in all experiments with  a momentum of $0.9$ and a weight decay of $5\times 10^{-4}$.
The plain VGG model is initialized with Kaiming initialization~\cite{he2015delving} and ResNets are initialized with Fixup~\cite{zhang2019fixup}.
We do not apply WA in the final classifier layer.
For classification, we train the models with data augmentation, random horizontal flipping and random crop, as in~\cite{zhong2017random}.
Further implementation details are given in each subsection.

\subsection{Experiments on CIFAR-10 \& CIFAR-100} \label{visu_WA}
\noindent \textbf{Comparing normalization methods.} We compare our proposed WeightAlign (WA) against various activation normalization methods, including Batch Normalization (BN)~\cite{ioffe2015batch}, Group Normalization (GN)~\cite{wu2018group}, Layer Normalization (LN)~\cite{ba2016layer}, Instance Normalization (IN)~\cite{ulyanov2016instance}. To demonstrate that WA is orthogonal to all these approaches, we also show the combination of WA with these activation normalization methods.

We conduct experiments on CIFAR-100 as shown in \tabl{normalization1}, where all models are trained with a  batch size of 64.
WeightAlign outperforms every normalization method except BN over large batch size.
By combining WeightAlign with sample statistics normalization methods, we can see that weightAlign adds additional information and  improves accuracy.  Especially, the GN+WA model achieves comparable performance of BN. 

To further show it's flexibility, we fine-tune a pre-trained BN model on CIFAR-100 with WA. It achieves 22.38\% error rate, which is the same as BN+WA model trained from scratch. This shows the possibility of finetuning WA on a pre-trained models with a sample-based normalization method.

On CIFAR-10 we do the same evaluation for ResNet50. BN and BN+WA are trained only with batch size of 64, and other methods are trained with batch size of 1 and 64. 
For batch size of 64 and 1, the initial learning rates are 0.01 and 0.001, respectively.  The baseline method~\cite{zhang2019fixup} is trained without any normalization methods.
Results are given in \tabl{normalization}. Because CIFAR-10 is less complex  than CIFAR-100, the performance gaps between different methods are thus less obvious. 
Nevertheless, our method is comparable to  other normalization methods and it allows a batch size of 1, which is not possible for BN. By combining WA with the sample-based normalization methods, we again observe that weight statistics normalization helps in addition to sample statistics normalization.

\begin{table}[t]
	\centering
	\footnotesize
	\caption{Error rate of ResNet50 on CIFAR-100 for classification. WA outperforms every normalization method except BN overlarge batch size. When combined with statistics normalization methods, WA improves accuracy. }
	\renewcommand{\arraystretch}{1.0}
  \setlength{\tabcolsep}{3.0mm}{
\begin{tabular}[c]{rlrl}
			\toprule
\multicolumn{4}{c}{CIFAR-100} \\ \cmidrule(lr){1-4}
Method & Error &  Method & error \\
			\midrule
			Baseline~\cite{zhang2019fixup} & 28.43 & WA  & 24.92  \\
			BN~\cite{ioffe2015batch}&  23.02  & BN + WA & 22.39    \\
	IN~\cite{ulyanov2016instance} & 25.58 &IN + WA &  24.63  \\		
LN~\cite{ba2016layer} & 26.78 & LN + WA & 24.01  \\
GN~\cite{wu2018group} & 25.46 & GN + WA &  23.64 \\
			\bottomrule
		\end{tabular}
		}\vspace{0.05in}

\label{table:normalization1}
\end{table}

\begin{table}[t]
	\centering
	\footnotesize
	\caption{Error rate of ResNet50 on CIFAR-10 for classification. Weight statistics normalization (WA) helps  in  addition  to  sample  statistics normalization. }
	\renewcommand{\arraystretch}{1.0}
  \setlength{\tabcolsep}{1mm}{
\begin{tabular}{rlrlrlrl}
			\toprule
\multicolumn{8}{c}{CIFAR-10} \\ \cmidrule(lr){1-8}
\multicolumn{4}{c}{Batch size 64} & \multicolumn{4}{c}{Batch size 1} \\
Method & Error & Method &  Error &  Method & Error & Method &  Error  \\
\cmidrule(lr){1-4} \cmidrule(lr){5-8}
			\midrule
            Baseline~\cite{zhang2019fixup} & 6.46 & WA & 6.21 & Baseline & 7.27 & WA & 6.61\\
            BN~\cite{ioffe2017batch} & 4.30 & BN+WA & 4.29 & BN & - & BN+WA & -\\
IN~\cite{ulyanov2016instance} & 6.49 & IN+WA & 6.42 & IN & 6.91 & IN+WA & 6.50\\
LN~\cite{ba2016layer} & 5.02 & LN+WA & 5.12 & LN & 6.82 & LN+WA & 5.76\\
GN~\cite{wu2018group} & 4.96 & GN+WA & 4.60 & GN & 5.79 & GN+WA & 5.51\\
			\bottomrule
		\end{tabular}
		}\vspace{0.05in}

			\label{table:normalization}
\end{table}

We also compare WA with Weight Normalization (WN)~\cite{salimans2016weight} on CIFAR-100 with VGG16 and ResNet50 networks. For VGG16, the error rates of WN, WA, GN are $6.8\%$, $4.8\%$and $7.4\%$ when the batch size is 1, and the ones of WN, WA, BN, GN are $8.6\%$, $5.4\%$, $5.8\%$ and $11.0\%$ when the batch size is 64. For ResNet50, WN cannot be trained. We can see WA and WN are workable regardless of the batch size for plain networks and WA outperforms WN in this model. For residual networks, WA can work alone without any activation normalization layer, while WN cannot. WN is only normalized by its norm to accelerate training. Without normalization layer in residual networks, WN cannot restrict the variance of the activation and the residual structure merges the activation from two branches, which leads to exploding gradients problem. Instead, WA normalizes the weights by zero-mean and a scaled standard deviation. This scale is determined by the size of convolutional filter and restricts the variance of weights and further maintains the variance of activations.\\

\begin{figure}[t]
  \centering
  \vspace{-2.5mm}
    \subfigure[VGG16]{
    \begin{minipage}[]{0.223\textwidth}
    \includegraphics[width=1\textwidth]{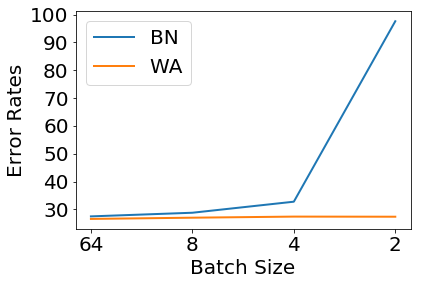}
    \end{minipage}
    }
    \subfigure[ResNet18]{
    \begin{minipage}[]{0.22\textwidth}
    \includegraphics[width=1\textwidth]{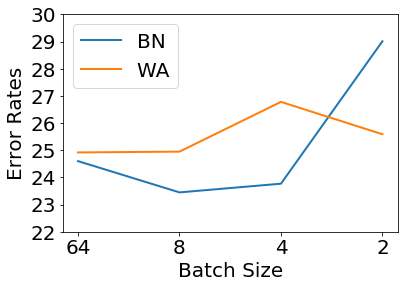}
    \end{minipage}
    }
\caption{Classification error \emph{vs.} batch sizes. The models are VGG16 and ResNet18 trained on CIFAR-100. The error rates of BN increase rapidly when small batch sizes are used. Our proposed method does not rely on the sample statistics and is independent of batch size dimension. The error rates of WA are stable over a wide range of batch sizes. }
	\label{fig:cifar_acc}
	\vspace{-3mm}
\end{figure}

\noindent \textbf{Small batch sizes.}
In \fig{cifar_acc}, we compare BN and WA with different batch sizes on VGG16 and ResNet18. We train with batch sizes of 64, 8, 4, 2 images per GPU. In all cases, the means and variances in the batch normalization layers are computed within each GPU and not synchronized. The x-axis shows different batch sizes and y-axis presents the validation error rates. From \fig{cifar_acc} (a), we observe that the performance of WA is comparable to that of BN on the plain network with large batches. The error rates of BN increase rapidly when reducing the batch size, especially with batch size smaller than 4. In contrast, our method focuses on the normalization of weight distributions which is independent of batch size. The error rates of our method are stable over different batch sizes. We can observe similar results in \fig{cifar_acc} (b) that our method performs stable over different batch size on residual network. It is interesting to see that the performance gap between WA and BN with ResNet18 is smaller than that with VGG16, which is due to the residual structure.\\

\noindent \textbf{Depth of residual networks.}
We evaluate depth of 18, 34, 50, 101, 152 for residual networks with batch sizes of 1, 64 per GPU. \tab{depth} shows the validation error rates on CIFAR-10.
We observe that WA is able to train very deep networks for both small and large batch sizes.\\

\begin{table}[t]
	\centering
	\footnotesize
	\caption{Applying WA to varying depth measured by error rate. WA  is  able  to  train  very  deep networks for both small and large batch sizes. }
	\renewcommand{\arraystretch}{1.0}
  \setlength{\tabcolsep}{3.5mm}{
\begin{tabular}[c]{rcc}
			\toprule
\multicolumn{3}{c}{CIFAR-10} \\ \cmidrule(lr){1-3} 
 &  Batchsize  1 &  Batchsize 64 \\
 Model (+WA)		& Error & Error \\
			\midrule
			ResNet18 & 5.65 & 6.23 \\
			ResNet34 & 5.84 & 5.78\\
	ResNet50 & 6.61 & 6.42  \\		
ResNet101 & 5.74 & 6.41 \\
ResNet152 & 6.09 & 6.52 \\
			\bottomrule
		\end{tabular}
		}
		\vspace{0.05in}

		\label{tab:depth}
\end{table}

\begin{figure*}[t]\vspace{-0.05in}
  \centering
  \subfigure[Baseline]{
  \begin{minipage}[t]{0.3\textwidth}
   \centering
    \includegraphics[width=1\textwidth]{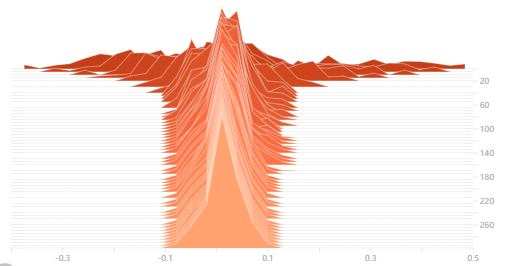}
    \end{minipage}
  }
  \subfigure[BN]{
  \begin{minipage}[t]{0.3\textwidth}
   \centering
    \includegraphics[width=1\textwidth]{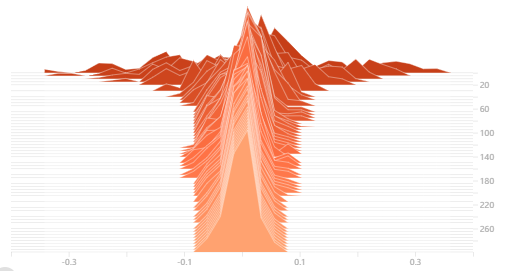}
    \end{minipage}
    }
    \subfigure[WA]{
  \begin{minipage}[t]{0.31\textwidth}
   \centering
    \includegraphics[width=1\textwidth]{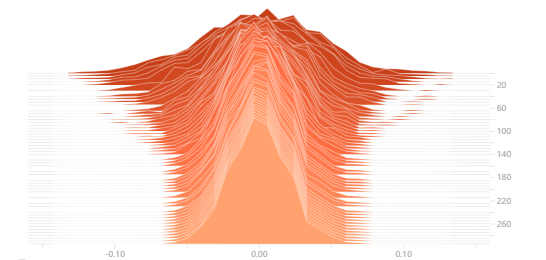}
    \end{minipage}
  
  }
  \caption{Channel weights distribution of trained ResNet50 versus epoch: (a) channel weights of baseline model. (b) channel weights of BN model. (c) channel weights of WA model. The x-axis represents the value of weights. The y-axis represents the epoch. From top to bottom, the epoch varies from 1 to 300. WA makes the distributions of channel weights symmetric around 0. With a symmetric distribution around zero for the weight, the activation can also have a symmetric distribution around zero.
   }
  \label{fig:weight_trained}
\end{figure*}

\noindent \textbf{Visualization of weights and activations in training.}
To verify the motivation, in this subsection, we visualize the
distribution of weights and activation by training a ResNet50 model on CIFAR-100.
A minibatch of 128 images from CIFAR-100 is input to the trained network to obtain features and weights.

\noindent \textbf{How are the filter weights distributed?} 
In \fig{weight_trained}, we visualize how the channel weight distribution changes with the epochs. 
\fig{weight_trained} (a) illustrates the baseline ResNet50 model  without using any normalization.
When using BN, as shown in \fig{weight_trained} (b), the weight distribution tends to be symmetric around zero. From \fig{weight_trained} (c), we note that the proposed WA method can realize the same functionality of BN making the weights symmetric and smooth during training. 
Channel and layer weights distribution during training of other normalization methods can be found in the supplementary material.
With a symmetric distribution around zero for the weight, the activation can also have a symmetric distribution around zero.

%

\begin{figure*}[t!]
 \centering
\begin{tabular}{cccccc}
\rotatebox{90}{\ \scriptsize{intermediate layer}}&\includegraphics[width=0.17\textwidth]{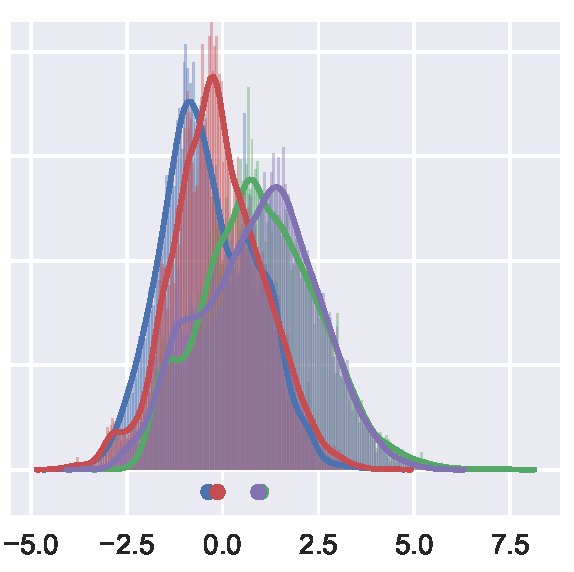} &
\includegraphics[width=0.17\textwidth]{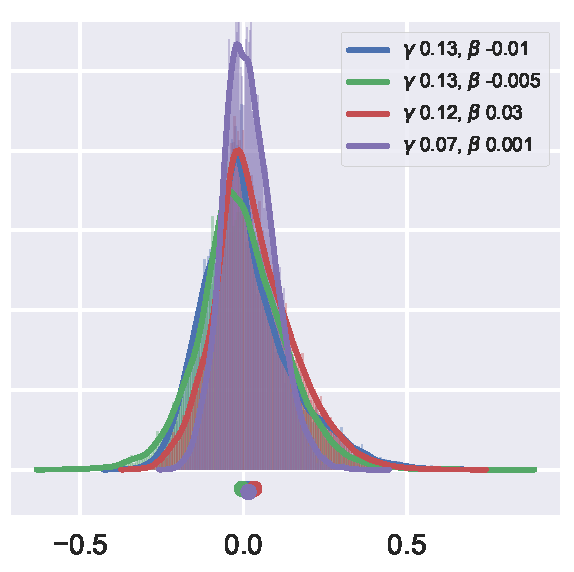} &
\includegraphics[width=0.17\textwidth]{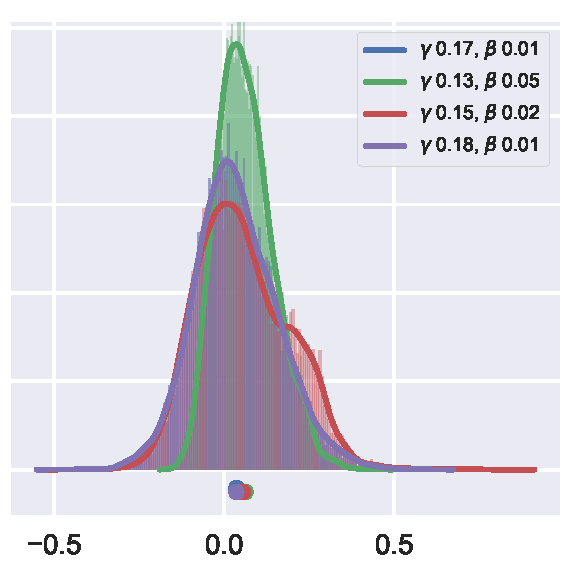} &
\includegraphics[width=0.17\textwidth]{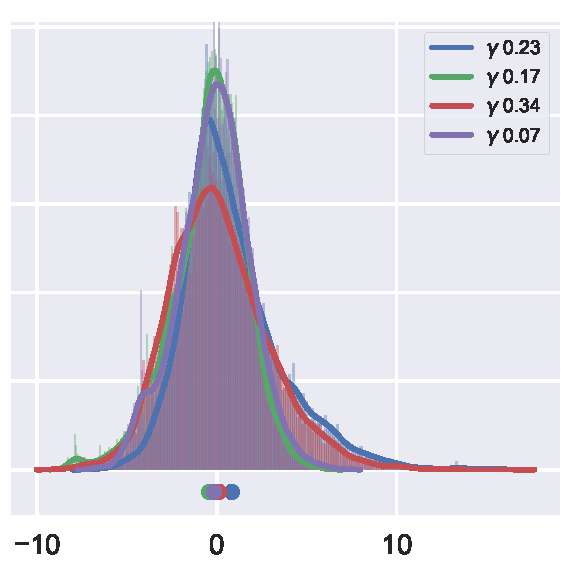}&
\includegraphics[width=0.17\textwidth]{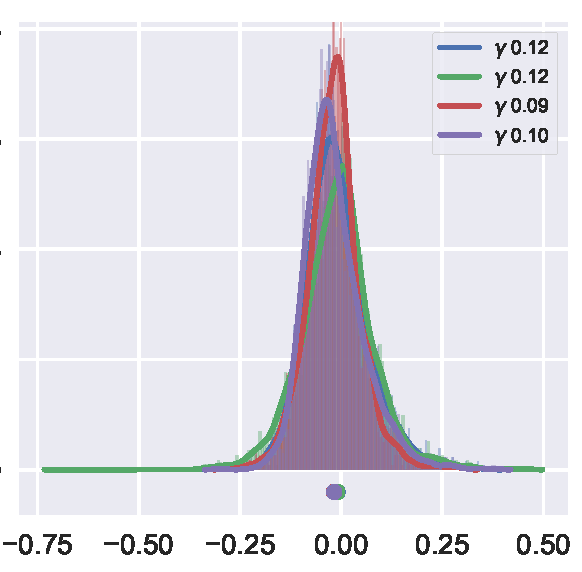} \\
\rotatebox{90}{\ \ \ \scriptsize{classifier layer}}&\includegraphics[width=0.17\textwidth]{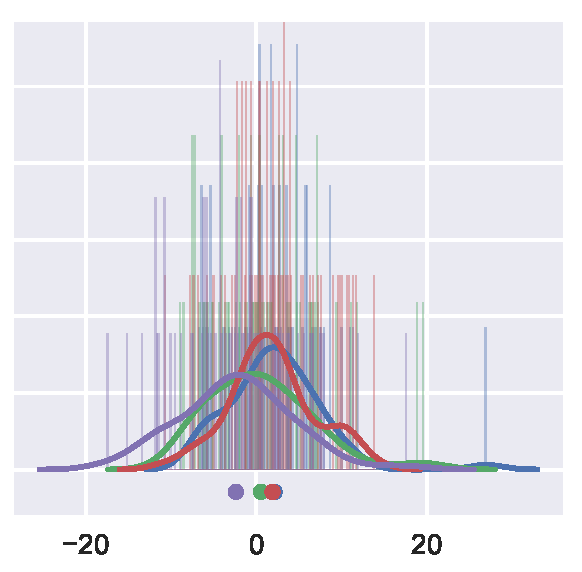} &
\includegraphics[width=0.17\textwidth]{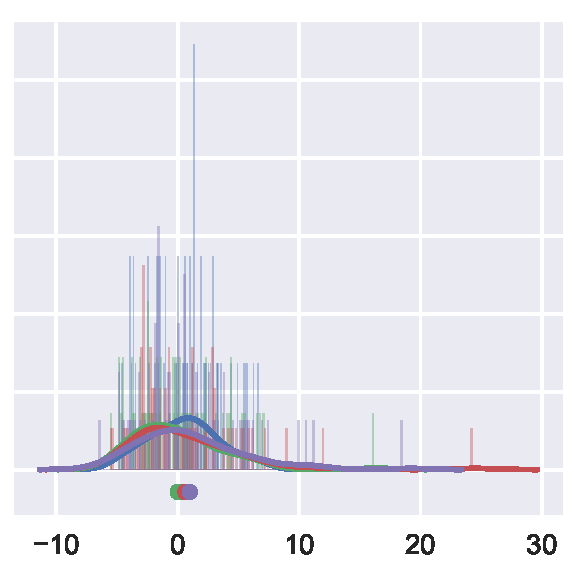} &
\includegraphics[width=0.17\textwidth]{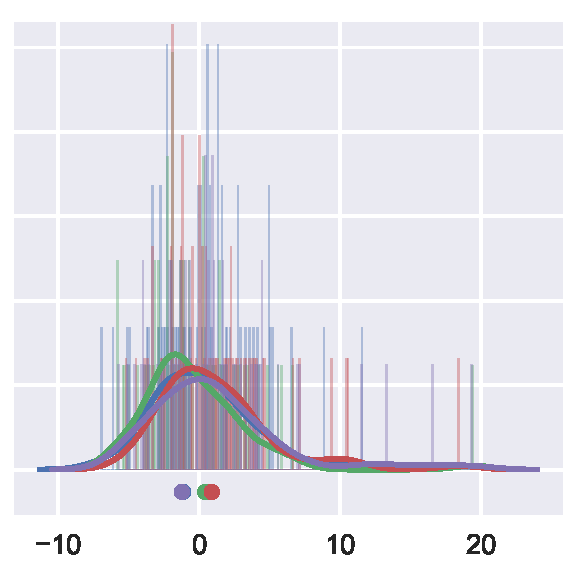} &
\includegraphics[width=0.17\textwidth]{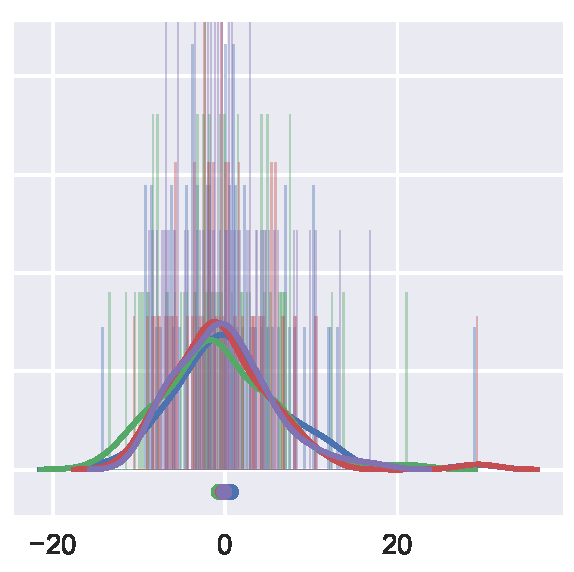}&
\includegraphics[width=0.17\textwidth]{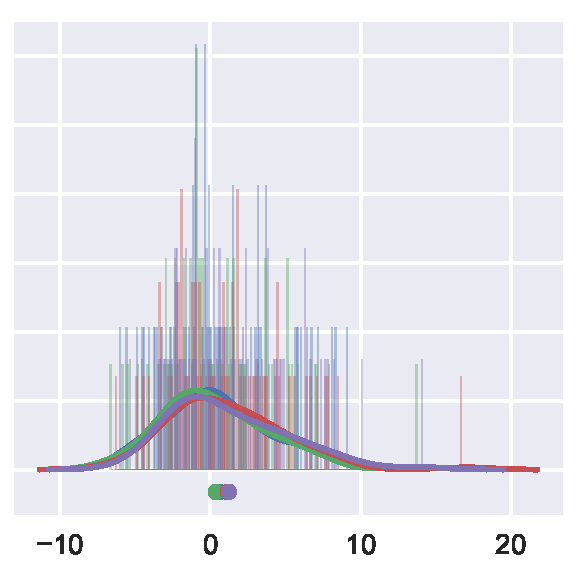} \\
&\scriptsize{(a) Baseline} &   \scriptsize{(b) BN} & \scriptsize{(c) GN }& \scriptsize{(d) WA} &  \scriptsize{(e) WA+GN} \\
\end{tabular}
\caption{Channel activation distributions of trained models on CIFAR-100. Different color represents different channel. $\gamma$ and $\beta$ are the parameters in each normalization. WA can align the channel activations by normalizing channel weights.}
 \label{fig:after_trainin}
\end{figure*}

\noindent \textbf{How are the channel activations distributed?} In~\fig{pSensitivity} we have found that before training, the proposed WA method can effectively normalize channel activations
by aligning the weights within each filter.  In~\fig{after_trainin} we further explore how the
channel activations distributed on a trained ResNet50 network.
 We visualize the activation distributions for 4 different channels taken from two layers, a middle layer (the 2rd conv layer of the 8th residual block) and the last classifier layer.
 Since the weight decay is used, the scale  and shift parameters $\gamma$, $\beta$ are converged to small values.
From~\fig{after_trainin} (d), we have similar observations as in \fig{pSensitivity} that our method can
normalize activation of  different channels to zero mean and same variance.
\fig{after_trainin} (e) also illustrates WA can be used in conjunction with GN to further normalize the activations.\\

\noindent \textbf{Functionality of different components.}
We align the filter weights by subtracting the mean and dividing by a properly scaled derivation. For our method, there are two components: $\mathbb{E}[w]  = 0 $ in~\eq{2mean} $\mathrm{Var}[w] = \frac{2}{n} $ in~\eq{con0}. We conduct an comparative results for our model with different components for VGG and ResNet18 models on CIFAR-100. In~\tab{centerlearning} we start with the baseline model~\cite{zhang2019fixup} and augment incrementally with the two components. We observe that both components contribute substantially to the performance of the whole model.\\

\begin{table}[t]
	\centering
	\caption{Comparative results for VGG16 and ResNet18 models with different components. Both components contribute substantially to the performance. }
	\renewcommand{\arraystretch}{1.2}
  \setlength{\tabcolsep}{2.5mm}{
	\begin{tabular}[c]{cccc}
		\toprule
		\multicolumn{2}{c}{Components in WA} & VGG16  & ResNet18 \\ 
		 $\mathbb{E}[w]  = 0 $ & $\mathrm{Var}[w] = \frac{2}{n} $  & Error & Error  \\ \midrule
		
		 $\times$ &$\times$ & - &  28.23   \\
		\checkmark  & $\times$  & 35.74 & 27.97 \\
	    $\times$ &\checkmark & - & 28.03 \\
		\checkmark &\checkmark & 27.85 & 24.92 \\
		\bottomrule
	\end{tabular}
	}
\vspace{0.05in}
	
	\label{tab:centerlearning}
		\vspace{-0.1in}
\end{table}

\subsection{Comparison with state-of-the-art}
\noindent \textbf{Image classification on ImageNet.}
We experiment with a regular batch size of 64 images per GPU on plain and residual networks. The total training epoch is 100. The initial learning rate is 0.1 and divided by 10 at the 30th, 60th and 90th epoch. \tab{imagenet} shows the top-1 and top-5 error rates of image classification on ImageNet.
\begin{table}[h!]
    \centering
    \caption{Top-1 and top-5 error rates of image classification on ImageNet. WA and WA+BN can work well on large and complex datasets. }
    \begin{tabular}{c c c}
    \toprule
     model & Top-1 (\%) Error & Top-5 (\%) Error\\ 
     \midrule
     VGG16$^*$ (Baseline) & 31.30 & 11.19\\
     VGG16 (BN)$^*$ & 29.58 & 10.16\\
     VGG16 (WA) & 29.78 & 10.23\\
     VGG16 (BN+WA) & \textbf{27.07} & \textbf{8.78}\\ \hline
     ResNet50 (Baseline)$^\dagger$~\cite{zhang2019fixup} & 27.60 & -\\
     ResNet50 (BN)$^*$ & 24.89 & 7.71\\
     ResNet50 (WA) & 26.62 & 8.91\\
     ResNet50 (BN+WA) & \textbf{24.04} & \textbf{7.12}\\
     \bottomrule
    \end{tabular}
    \begin{flushleft}
    The $^*$ denotes that we directly test the PyTorch pretrained model.\\
    The $^\dagger$ denotes the numbers from the reference.\\
    \end{flushleft}
    
    
    \label{tab:imagenet}
    \vspace{-0.1in}
\end{table}
For plain networks, WA performs relatively well as BN, only 0.2\% difference on top-1 error rate. Combining WA with BN outperforms BN on VGG16 by 2.51\% top-1 error rate and 1.38\% top-5 error rate. 
For residual networks, WA performs slightly worse than BN, but combined BN+WA surpasses BN by 0.85\% on top-1 error rate and 0.59\% on top-5 error rate.
This shows that WA and WA+BN can work well on large and complex datasets.\\


\noindent \textbf{Domain adaptation on Office-31.}
 \tab{domain} presents overall scores of different normalization methods and their combination with WA on Office-31 dataset.
Here we use ResNet50 models pretraied on CIFAR-100, which are obtained from previous experiments.
To perform domain adaptation experiments, we finetune the pretrained model on a source domain, and validate it on two target domains with 0.001 initial learning rate, e.g. finetune on A, and validate on W and D.
Model with WA consistently outperforms model with GN, IN and LN.
\tab{domain} also shows that our method WA can be applied to other normalization methods successfully and improves domain adaptation performance.
In Office-31, the three domains A, W and D have different sample distribution.
Thus the normalization methods that rely on sample statistics perform worse than our method WA, which only applies on weights and is independent of sample statistics.
When combined with WA, other activation normalization methods can benefit from the sample-independent merit of WA, which boosts their domain adaptation performance. \\

\begin{table}[!t]
	\centering
	\footnotesize
	\caption{Classification accuracies (\%) on Office-31 dataset (ResNet50). WA can be applied to other normalization methods and improves domain adaptation performance. }
    \setlength{\tabcolsep}{0.9mm}{
		\begin{tabular}[c]{lllllllll}
			\toprule
			Method & A $ \rightarrow$ W & W $ \rightarrow$ A& A $ \rightarrow$ D  & D $ \rightarrow$ A & W $ \rightarrow$ D  & D $ \rightarrow$ W & Avg \\ \rowcolor{gray!15}
			\midrule
WA & 34.62 & 42.89 & 52.05 & \textbf{44.09} & 87.67 & 80.00 & 56.89 \\ 
BN~\cite{ioffe2015batch} & 46.92 & 48.91 & 57.53 & 37.11 & 90.41 & 70.00 & 58.48 \\ \rowcolor{gray!15}
BN + WA & 49.23 & 36.38 & \textbf{58.90} & 40.00 & \textbf{91.78} & 76.92 & 58.87 \\
GN~\cite{wu2018group} & 46.92 & 42.89 & 50.68 & 39.76 & 82.19 & 75.38 & 56.30 \\ \rowcolor{gray!15}
GN + WA & 46.15 & \textbf{54.46} & 50.68 & 37.83 & 87.67 & 78.46 & 59.21 \\ 
IN~\cite{ulyanov2016instance} & 36.92 & 33.98 & 41.10 & 37.11 & 84.93 & 76.92 & 51.83 \\ \rowcolor{gray!15}
IN + WA & 39.23 & 41.45 & 38.36 & 38.07 & 90.41 & 73.85 & 53.56 \\ 
LN~\cite{ba2016layer} & 37.69 & 29.89 & 45.20 & 33.97 & 79.45 & 69.23 & 49.24 \\ \rowcolor{gray!15}
LN + WA & \textbf{50.77} & 42.89 & 56.16 & 41.69 & \textbf{91.78} & \textbf{80.77} & \textbf{60.68} \\
			\bottomrule
		\end{tabular}
	}
	\vspace{0.05in}
	
	\label{tab:domain}
	\vspace{-0.1in}
\end{table}

\noindent \textbf{Semantic segmentation on PASCAL VOC 2012.}
To investigate the effect on small batch size, we conduct experiments on semantic segmentation with PASCAL VOC 2012 dataset. We select the DeepLabv3~\cite{chen2018encoder} as baseline model in which the ResNet50 pre-trained on ImageNet is used as backbone. Given the pretrained ResNet50 models with BN and BN+WA on ImageNet, we finetune them on PASCAL VOC 2012 for 50 epochs with batch size 4, 0.007 learning rate with polynomial decay and multi-grid (1,1,1), and evaluate the performance on the validation set with output stride 16. The evaluation mIoU of BN method is 73.80\%, and that of BN+WA is 74.87\%. Combined with WA, the original BN performance is improved by 1.03\%. This further shows that WA improves other normalization methods.\\


\section{Conclusion}
We propose WeightAlign; a method that re-parameterizes the weights by the mean and scaled standard derivation computed within a filter.
We experimentally demonstrate WeightAlign on five different datasets.
WeightAlign does not rely on the sample statistics, and performs on par with Batch Normalization regardless of batch size.
WeightAlign can also be combined with other activation normalization methods and consistently improves on Batch Normalization, Group Normalization, Layer Normalization and Instance Normalization on various tasks, such as image classification, domain adaptation and semantic segmentation.

\bibliographystyle{IEEEtran}
\bibliography{egbib}
\clearpage
\section*{Appendix}
\section{Proof of symmetric}
Given two independent random variable $X$ and $Y$, the distribution of product random variable $Z$, where $Z=XY$, can be found as follows,
\begin{equation}
    f_Z(z) = \int_{-\infty}^{\infty}\frac{1}{\left|t\right|}f_X(t)f_Y(\frac{z}{t})\,dt.
\end{equation}
If the distribution of $X$ is continuous at 0, then we have,
\begin{equation}
    \begin{aligned}
        P(Z\leq z) &= P(XY\leq z)\\
    &= P(Y\leq\frac{z}{X}|X>0)P(X>0)+P(Y\geq\frac{z}{X}|X<0)P(X<0)\\
    &= \int_0^{\infty}P(Y\leq\frac{z}{t})f_X(t)\, dt+\int^0_{-\infty}P(Y\geq\frac{z}{t})f_X(t)\, dt.
    \end{aligned}
\end{equation}
We take the derivation of both sides w.r.t. z and we get,
\begin{equation}
    \begin{aligned}
    f_Z(z) &= \int_0^{\infty}\frac{1}{t}f_Y(\frac{z}{t})f_X(t)\, dt + \int_{-\infty}^0\frac{-1}{t}f_Y(\frac{z}{t})f_X(t)\, dt\\
    &= \int_{-\infty}^{\infty}\frac{1}{\left|t\right|}f_X(t)f_Y(\frac{z}{t})\, dt.
    \end{aligned}
\end{equation}
Suppose the distribution of random variable $Y$ is symmetric around zero, where $f_Y(y) = f_Y(-y)$. Then we can have,
\begin{equation}
    \begin{aligned}
    f_Z(z) &= \int_{-\infty}^{\infty}\frac{1}{\left|t\right|}f_X(t)f_Y(\frac{z}{t})\, dt \\
    &= \int_{-\infty}^{\infty}\frac{1}{\left|t\right|}f_X(t)f_Y(\frac{-z}{t})\, dt \\
    &= f_Z(-z).
    \end{aligned}
\end{equation}
Therefore, the distribution of product random variable $Z$ will be symmetric around zero, if the distribution of one of independent random variable $X$ and $Y$ is symmetric around zero.

\section{Explanation of WA equation form}
In section \ref{weightalignEq}, we give out the expression for WA as  \eq{reparmeterize1}.
Compared with BN, WA does not have the $\beta$ term since it manipulates the weight directly instead of the activations.
If we add a $\beta$ term to \eq{reparmeterize1}, it will result in,
 \begin{equation}
\hat{{w}} = \gamma \ \frac{{w}- \mathbb{E}[{w}]}{\sqrt{{n}/{2} \cdot \mathrm{Var}[{w}]} } + \beta.
\label{eq:reparmeterize1-beta}
\end{equation}
When it multiplies with the input activations $\mathbf{x}$, we will get an extra $\beta\mathbf{x}$, which is similar to the element in residual blocks.
This will introduce the drift of activation variance again, which goes against our original intention.

\section{Additional Experiments and Visualization}

\subsection{Scale factor in WeightAlign} 
\begin{figure}[htbp]
    \centering
    \includegraphics[width=0.5\textwidth]{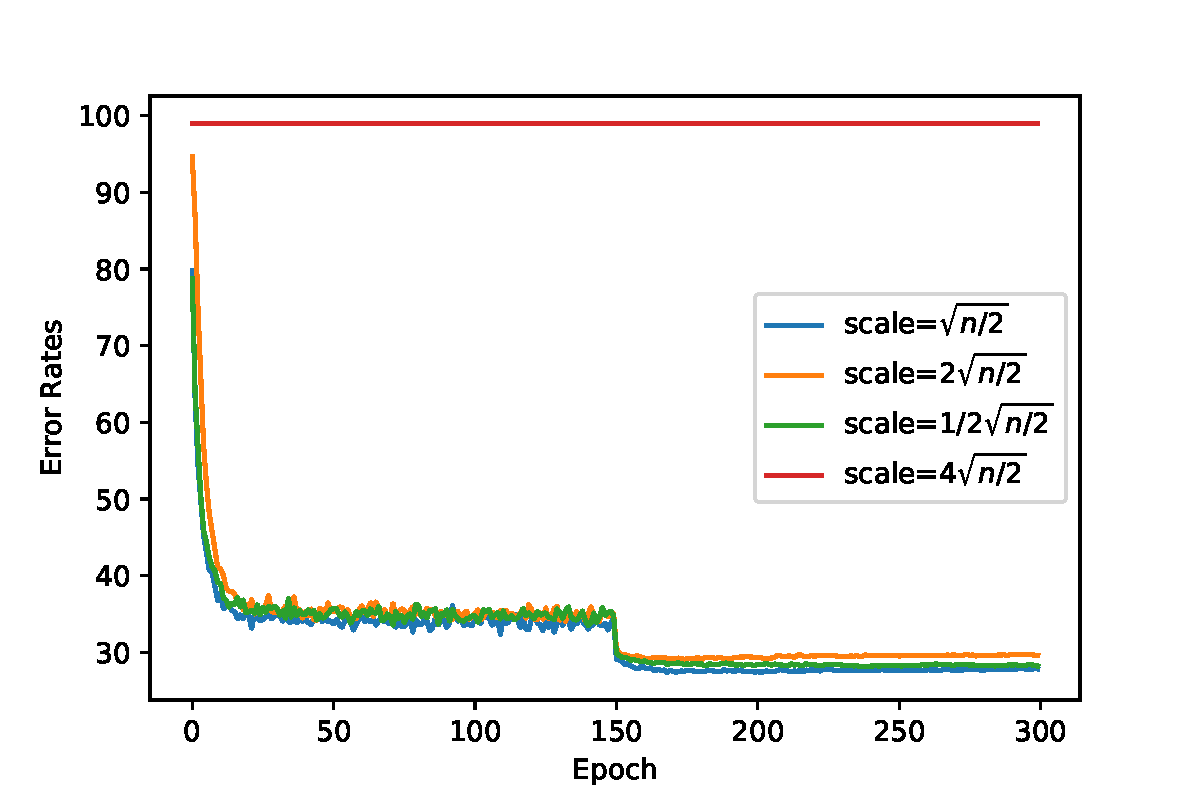}
    \caption{
    Ablation study of scale factor. Validation error rates with different scale factors on CIFAR-100 (ResNet18). A slightly different scale factor can cause a failure of training like the red line. A proper scale factor is important to stabilize the optimization. Our scale achieves the best performance among other trainable scale factors.}
    \vspace{-2mm}
    \label{fig:ablation}
    \vspace{-0.05in}
\end{figure}

We here conduct an ablation study experiments to validate the effect of our scale factor $\sqrt{n/2}$ in \eq{reparmeterize1}.
 Specially, we compare our scale factor  with  a
 $0.2\mathrm{x}$ our scale factor, $2\mathrm{x}$ our scale factor and $4\mathrm{x}$ our scale factor.
 The experiments are shown in \fig{ablation}.
We find that a similar scale to $\sqrt{n/2}$ will lead to similar performance. But our scale $\sqrt{n/2}$ has the best performance. Any other scale factors would cause the failure of training. Thus a proper scaled derivation plays an important role to make the training stable.

\begin{figure*}[htb]
 \centering
\begin{tabular}{rc}
\rotatebox{90}{\ \ \ \  {(a) Baseline}} &
\includegraphics[width=0.95\textwidth]{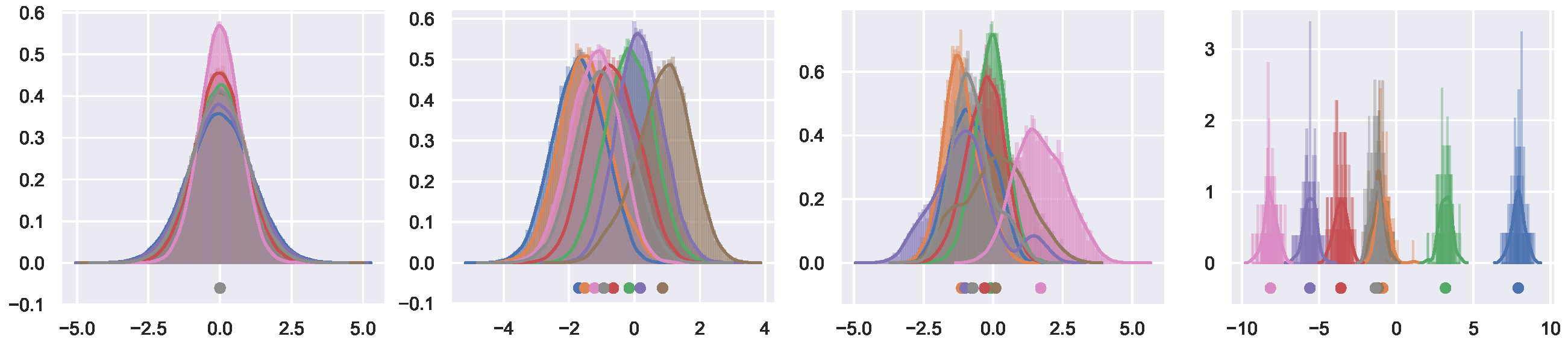}  \\
\rotatebox{90}{\ \ \ \ \ \ \   {(b) WA}} &
\includegraphics[width=0.95\textwidth]{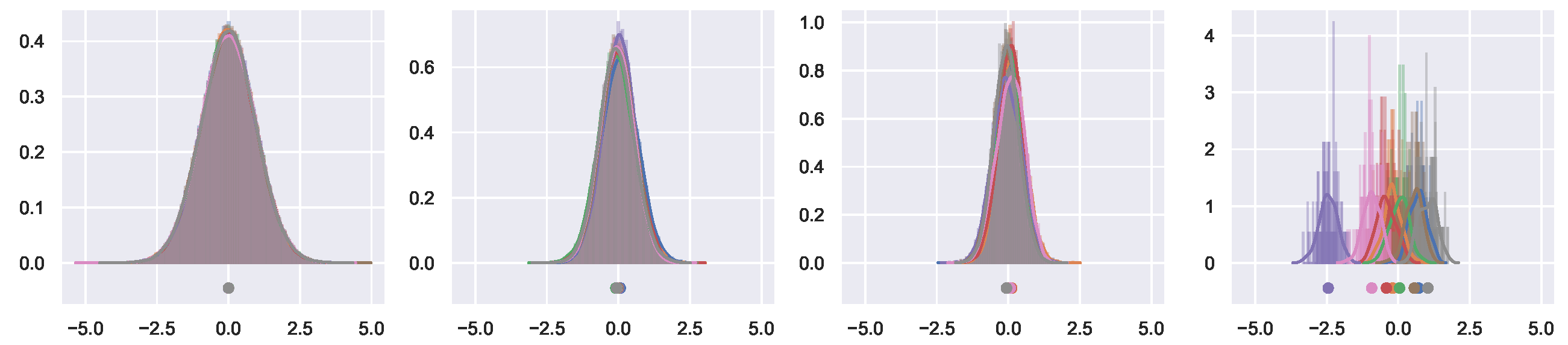}  \\
\rotatebox{90}{\ \ \ \ \ \ \   {(c) BN}} &
\includegraphics[width=0.95\textwidth]{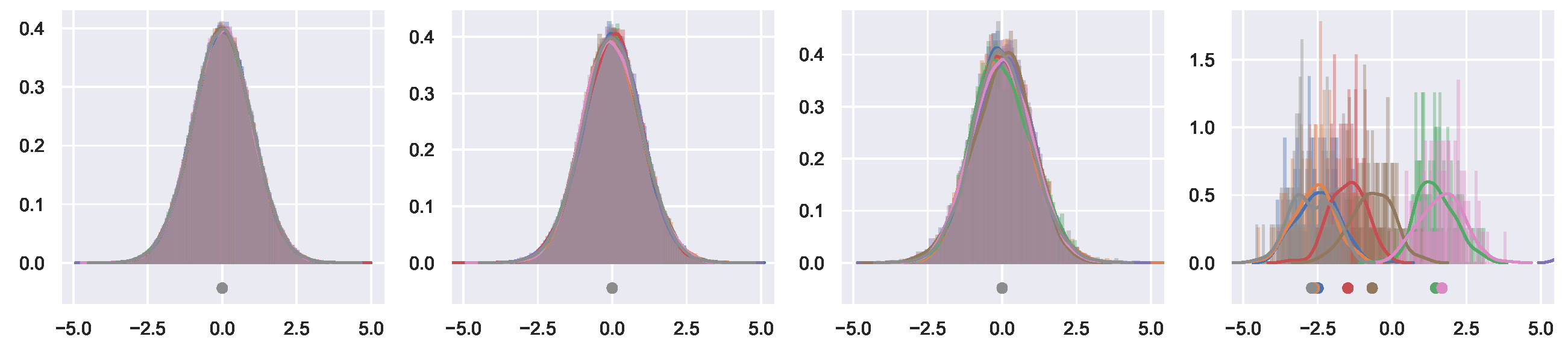}  \\
\rotatebox{90}{\ \ \ \ \ \ \   {(d) LN}} &
\includegraphics[width=0.95\textwidth]{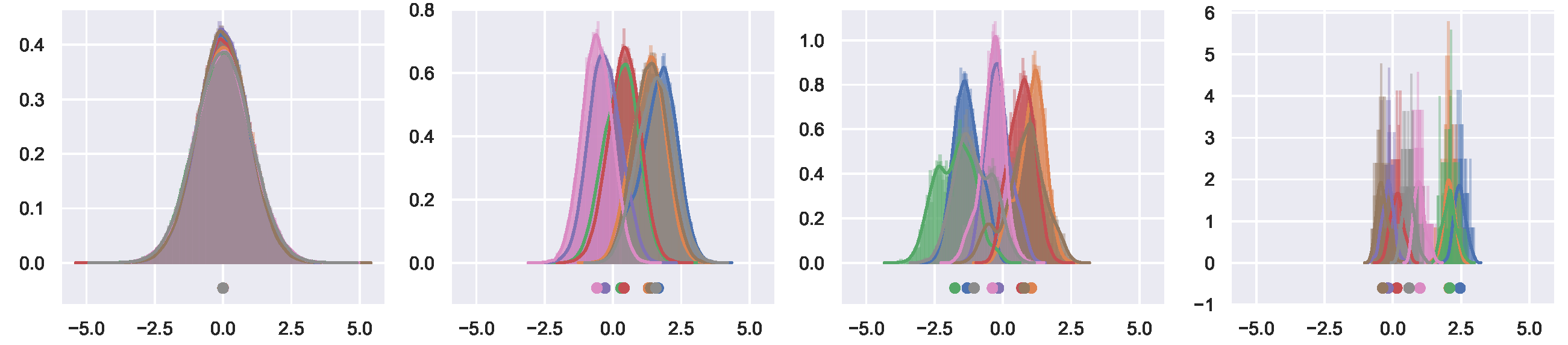}  \\
\rotatebox{90}{\ \ \ \  \ \ \  {(e) IN}} &
\includegraphics[width=0.95\textwidth]{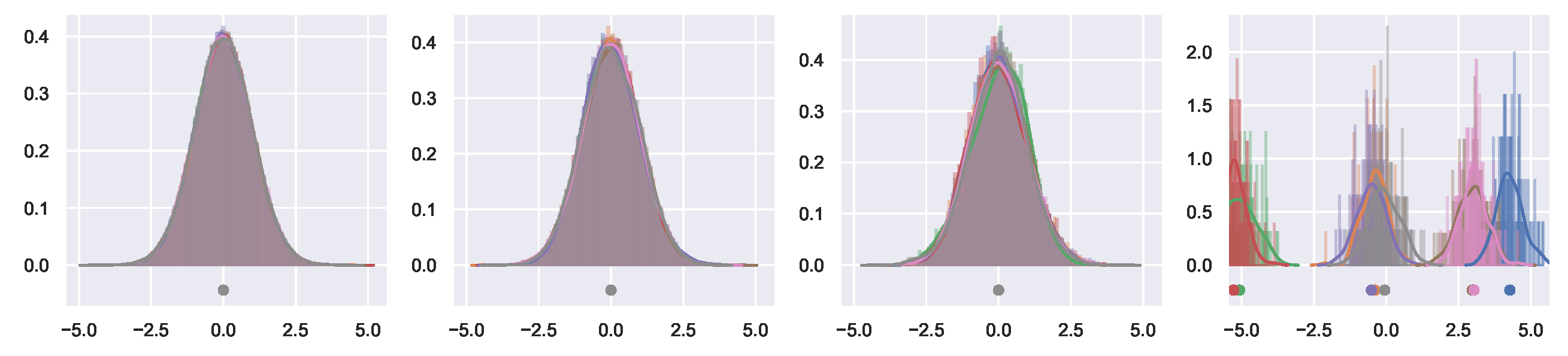} 
\\
\rotatebox{90}{\ \ \ \  \ \ \ {(f) GN}} &
\includegraphics[width=0.95\textwidth]{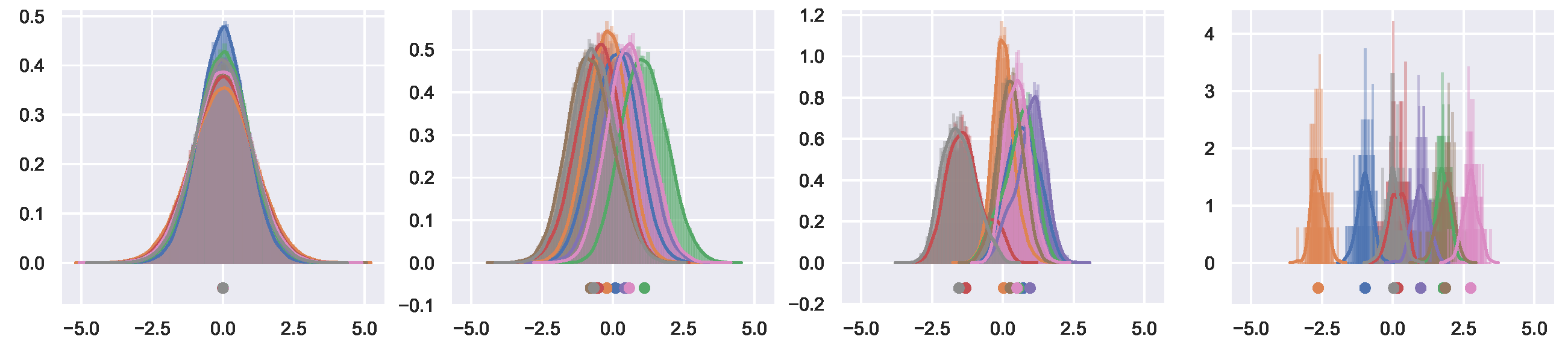}  \\
\end{tabular}
\caption{
Each color represents the activation distribution of different channels for different layers.
 The first three columns denote 1st, 3rd, 7th convolutional layers and the last one
presents the last classification layer before softmax.
All normalization methods can reduce internal covariate shift to some extent comparing with baseline.}
 \label{fig:app1}
\end{figure*}

\subsection{Empirical analysis in Section~\ref{Analysis}}
\begin{figure*}[htb]
 \centering
\begin{tabular}{rc}
\rotatebox{90}{\ \ \ \  {(a) BN+WA}} &
\includegraphics[width=0.95\textwidth]{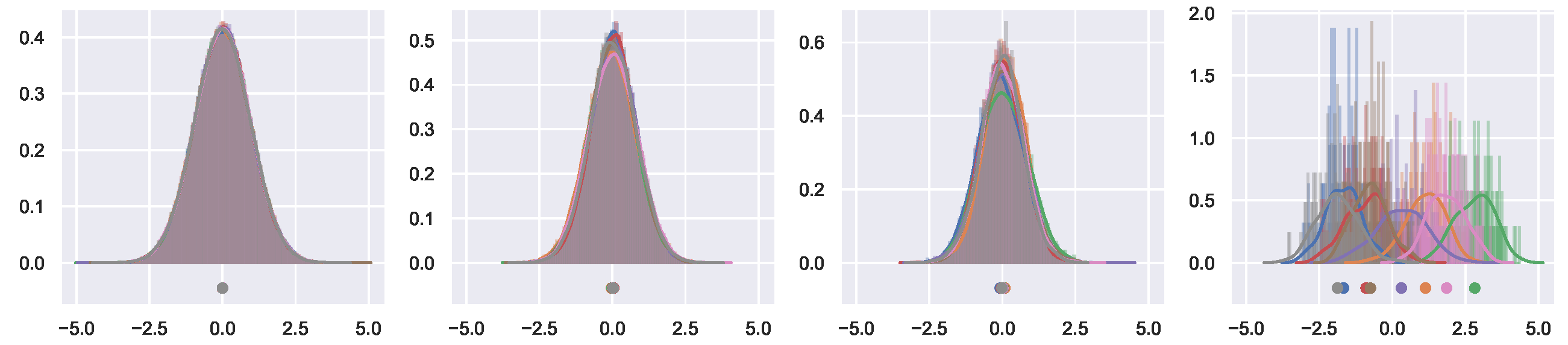}  \\
\rotatebox{90}{\ \ \ \ {(b) LN+WA}} &
\includegraphics[width=0.95\textwidth]{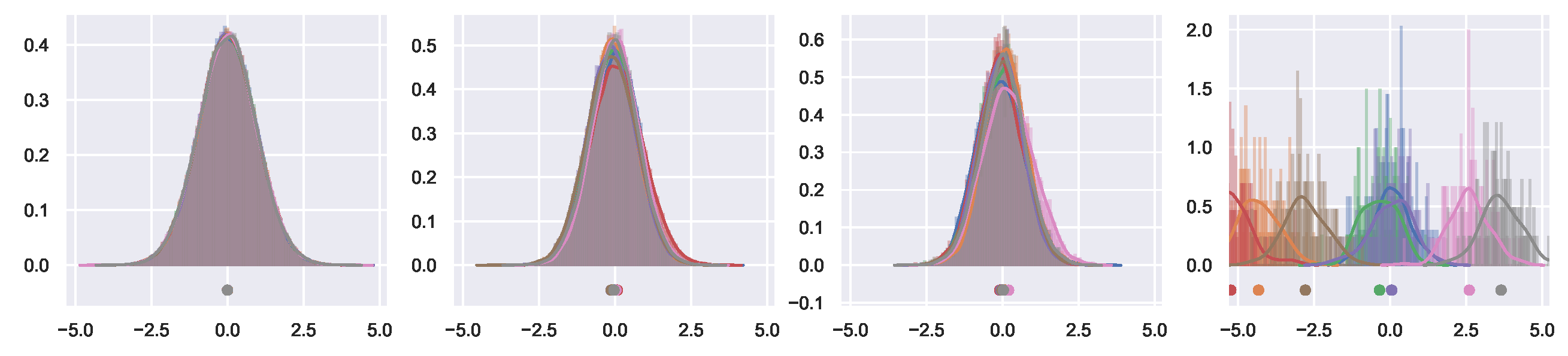}  \\
\rotatebox{90}{\ \ \ \  {(c) IN+WA}} &
\includegraphics[width=0.95\textwidth]{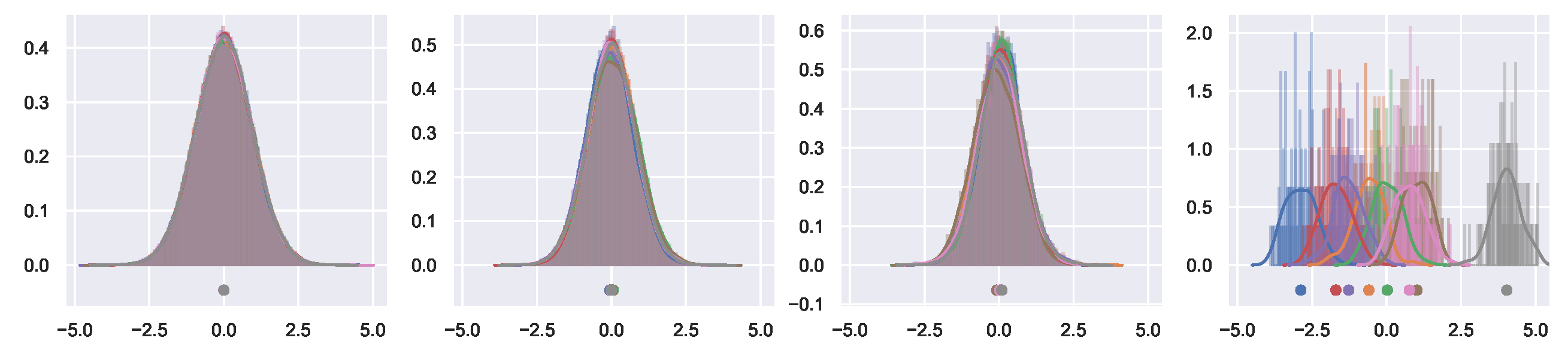}  \\
\rotatebox{90}{\ \ \ \ {(d) GN+WA}} &
\includegraphics[width=0.95\textwidth]{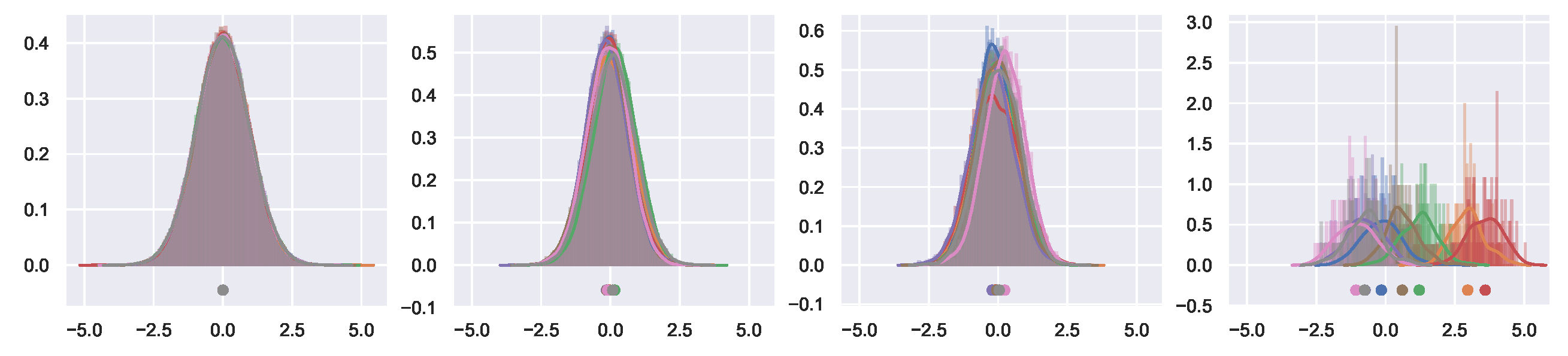}  \\
\end{tabular}
\caption{ The proposed WA can be used in conjunction with BN, LN, IN and
GN. The first three columns denote 1st, 3rd, 7th convolutional layers and the last one
presents the last classification layer before softmax. Note that the activation we plot here is before passing through specific normalization layer.}
 \label{fig:app3}
\end{figure*}

We further show activation distributions for different normalization methods over 8 different channels taken from four different layers: the 1st, 3rd, 7th intermediate convolutional layers and the last classification layer.
\fig{app1} shows the comparison between baseline and other normalization methods including WA.
\fig{app3} shows the cases when IN, LN and GN are used in conjunction with our WA.


\subsection{Visualizations of weights and activations in Section~\ref{visu_WA}}

\begin{figure*}[htb]
 \centering
\begin{tabular}{ccc}
\includegraphics[width=0.33\textwidth]{FigV/fixup_none_channel_weight.png} & \includegraphics[width=0.33\textwidth]{new_fig/layer_5_channel_4.png}  &\includegraphics[width=0.33\textwidth]{FigV/fixup_bn_channel_weight.png} \\
		\scriptsize{(a) Baseline}  & 
		\scriptsize{(b) WA } &  \scriptsize{(c) BN } \\
\includegraphics[width=0.33\textwidth]{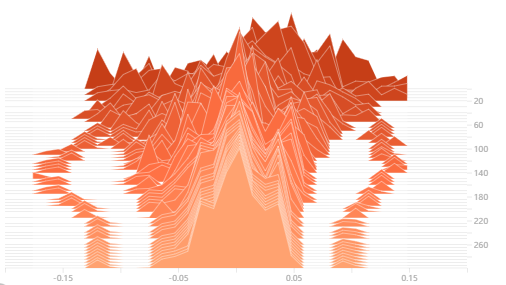} & \includegraphics[width=0.33\textwidth]{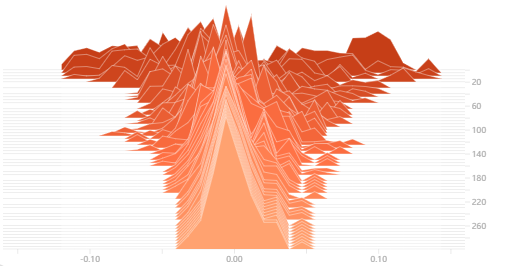}  &\includegraphics[width=0.33\textwidth]{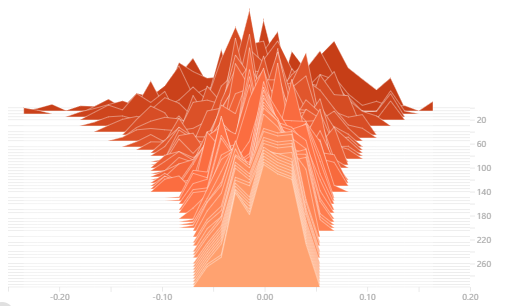} \\
		\scriptsize{(d) LN}  & 
		\scriptsize{(e) IN } &  \scriptsize{(f) GN } \\
\end{tabular}
\caption{Weights distributions of a single channel in a convolutional layer for different normalization methods. WA and BN have smooth and symmetric weight distributions.}
 \label{fig:Chaweight1}
\end{figure*}

\begin{figure*}[htb]
 \centering
\begin{tabular}{ccc}
\includegraphics[width=0.33\textwidth]{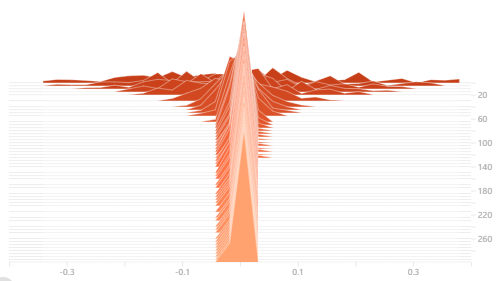} & \includegraphics[width=0.33\textwidth]{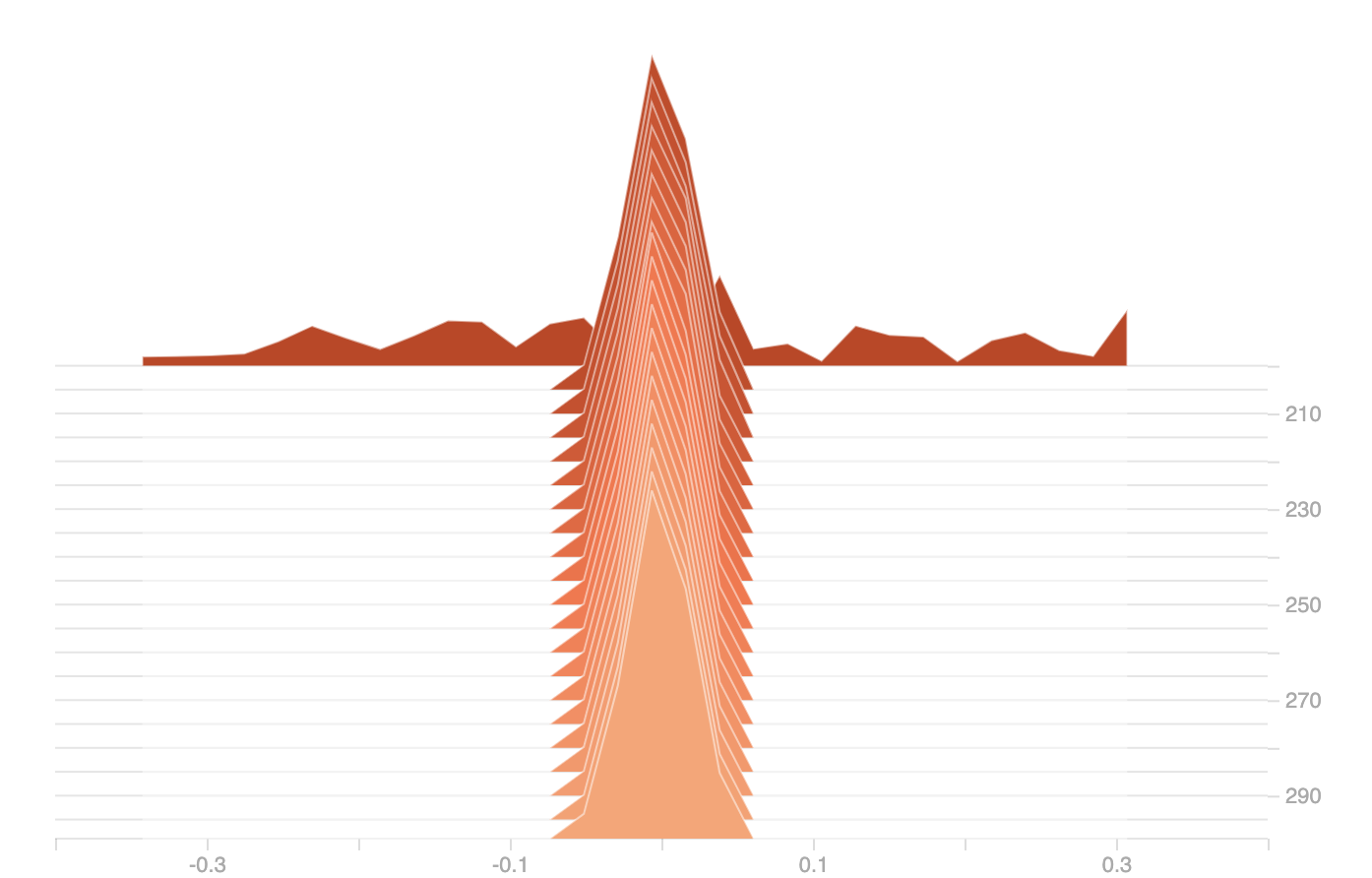}  \\
		\scriptsize{(a) WA + BN }  & 
		\scriptsize{(b) WA + LN } \\ 
		\includegraphics[width=0.33\textwidth]{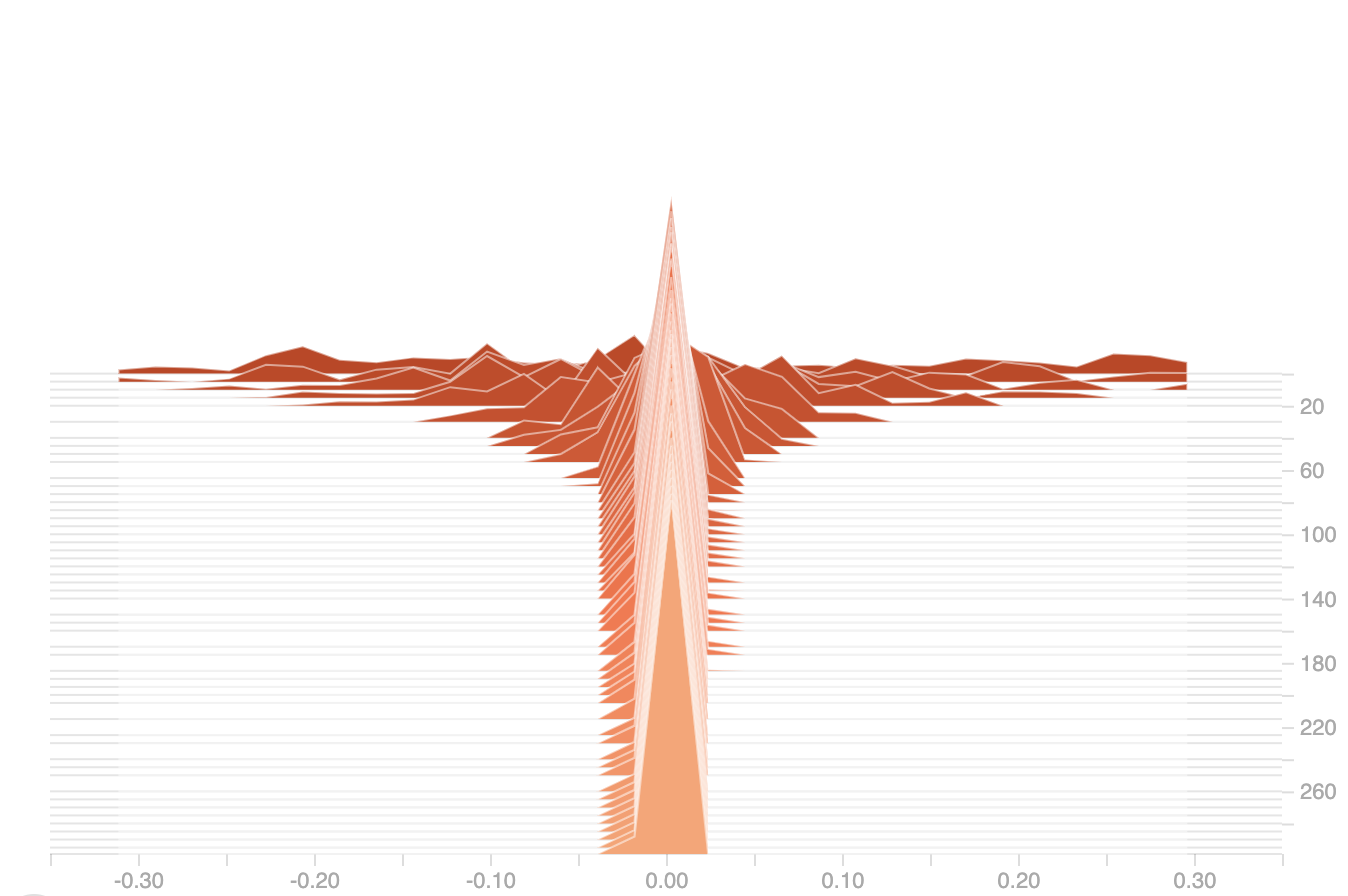} 
		&
\includegraphics[width=0.33\textwidth]{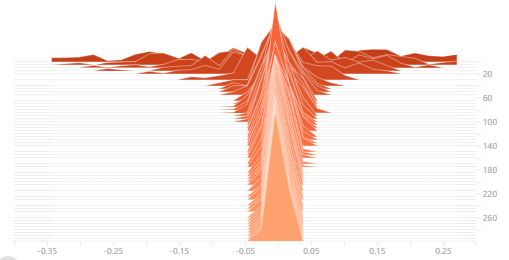} \\
\scriptsize{(c) WA + IN } & \scriptsize{(d) WA + GN }  \\
\end{tabular}
\caption{Weights distributions of a single channel in a convolutional layer for different normalization methods in conjunction with WA method. Adding WA smooths and symmetrizes the weight distributions.}
 \label{fig:Chaweight2}
\end{figure*}

\begin{figure*}[htb]
 \centering
\begin{tabular}{ccc}
\includegraphics[width=0.33\textwidth]{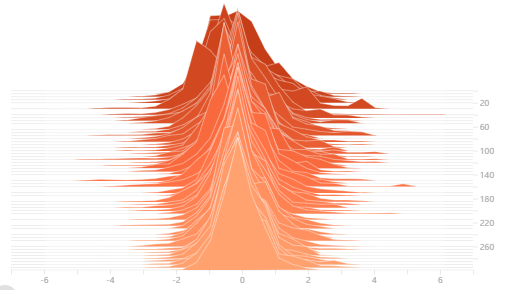} & \includegraphics[width=0.33\textwidth]{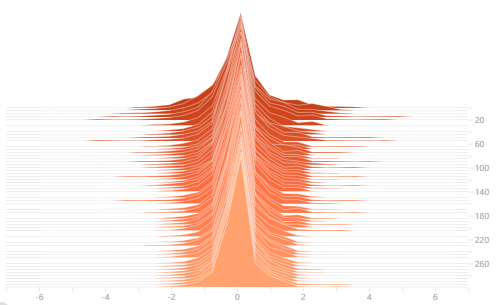}  &\includegraphics[width=0.33\textwidth]{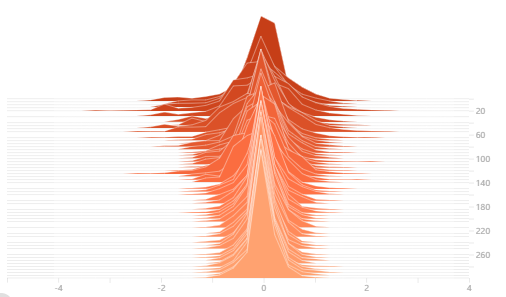} \\
		\scriptsize{(a) Baseline}  & 
		\scriptsize{(b) WA } &  \scriptsize{(c) BN } \\
\includegraphics[width=0.33\textwidth]{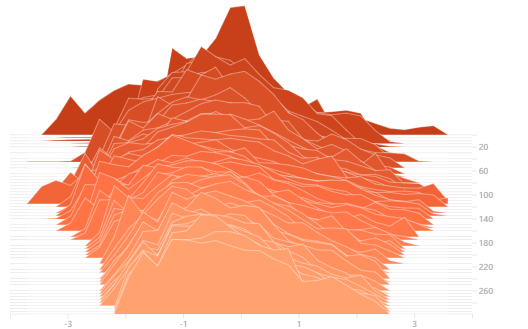} & \includegraphics[width=0.33\textwidth]{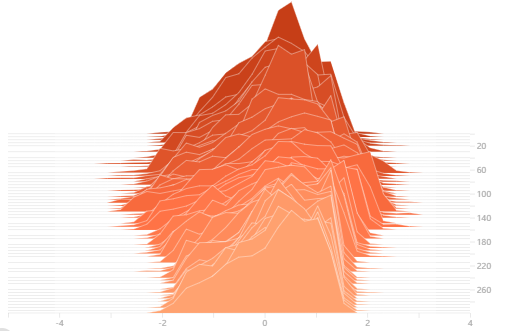}  &\includegraphics[width=0.33\textwidth]{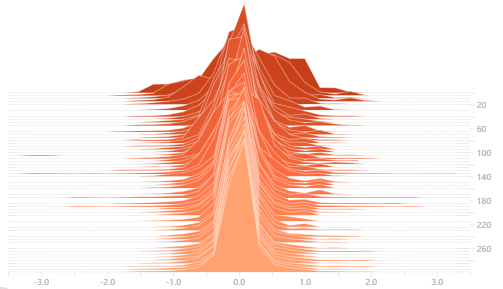} \\
		\scriptsize{(d) LN}  & 
		\scriptsize{(e) IN } &  \scriptsize{(f) GN } \\
\end{tabular}
\caption{Activation distributions of a single channel in a convolutional layer for different normalization methods. WA and BN have smooth and symmetric weight distributions.}
 \label{fig:Chaactivation1}
\end{figure*}

\begin{figure*}[htb]
 \centering
\begin{tabular}{cc}
\includegraphics[width=0.33\textwidth]{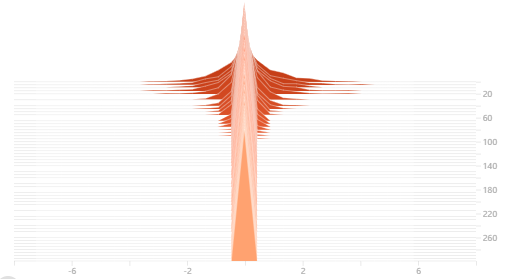} & \includegraphics[width=0.33\textwidth]{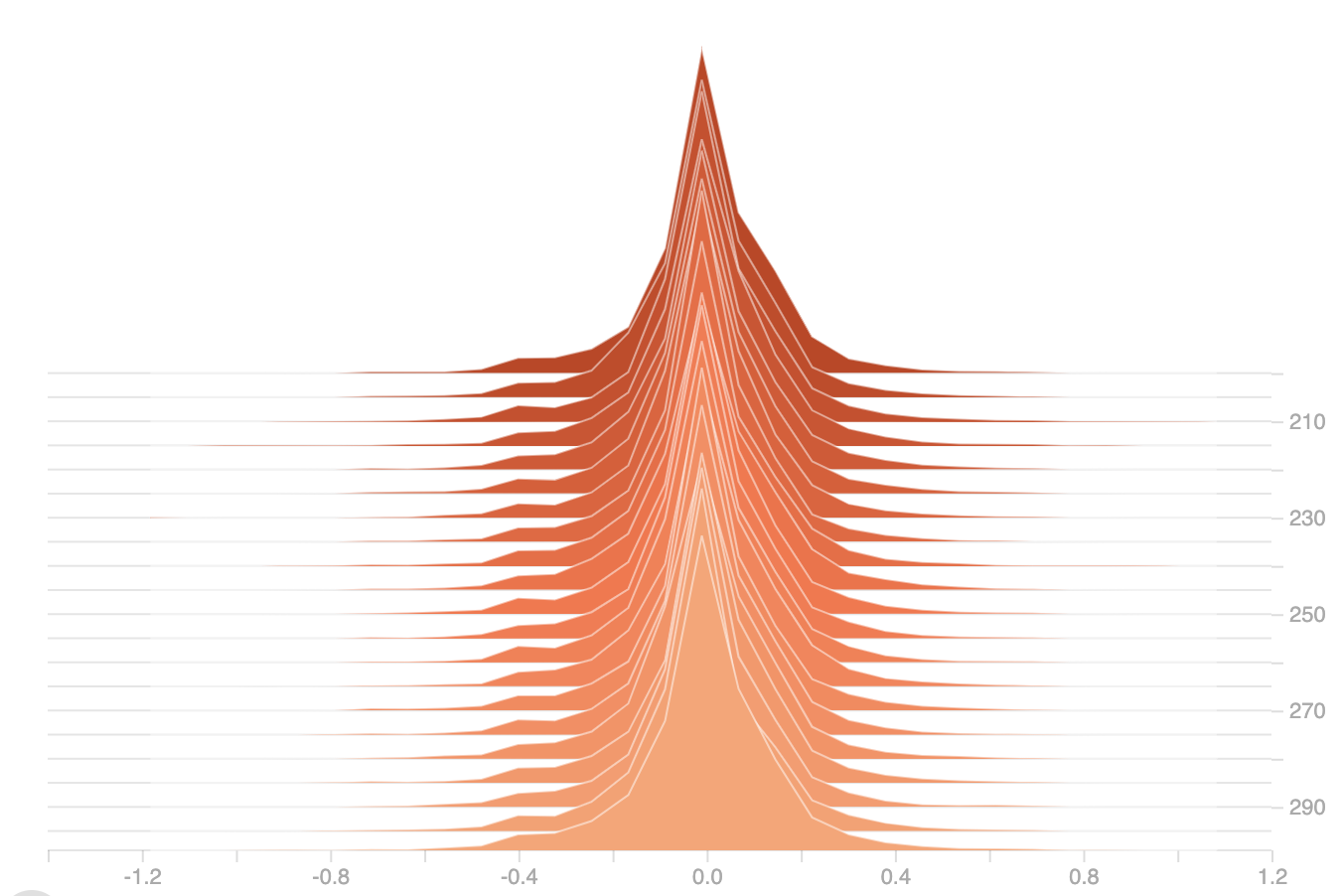} \\
		\scriptsize{(a) WA + BN }  & 
		\scriptsize{(b) WA + LN } \\ 
\includegraphics[width=0.33\textwidth]{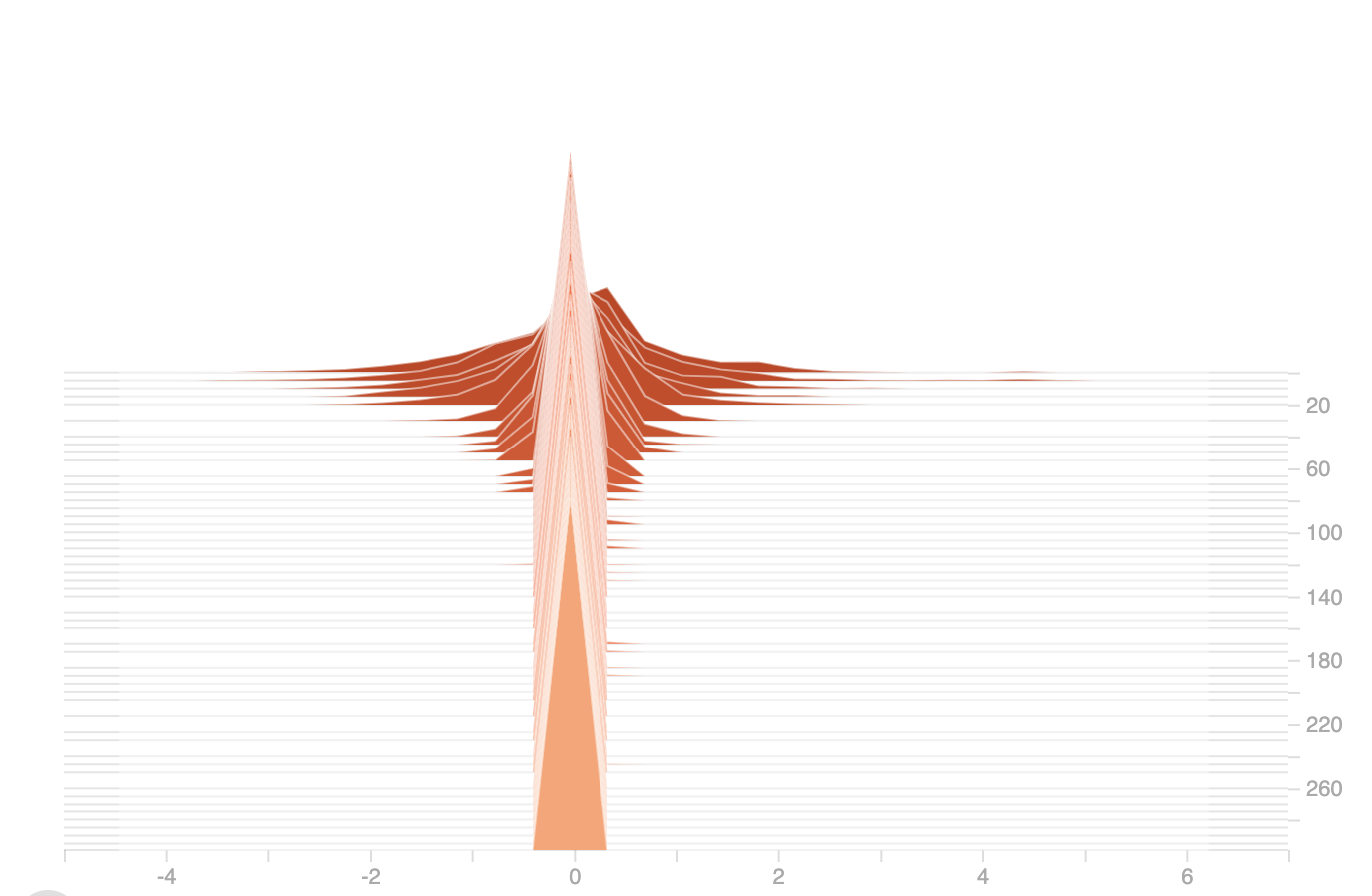}&\includegraphics[width=0.33\textwidth]{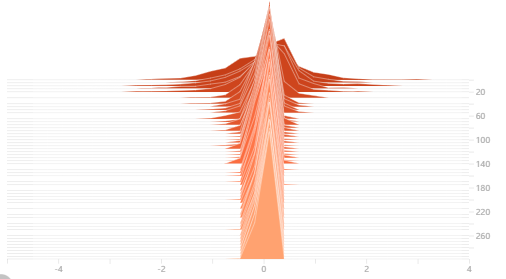}\\
\scriptsize{(c) WA + IN }&\scriptsize{(d) WA + GN } \\
\end{tabular}
\caption{Activation distributions of a single channel in a convolutional layer for different normalization methods in conjunction with WA method. Adding WA smooths and symmetrizes the activation distributions.}
 \label{fig:Chaactivation2}
\end{figure*}

We visualize the distributions of weights and activations as epoch increases during training. \fig{Chaweight1} and \fig{Chaweight2} show weights distributions of a single channel in a convolutional layer. The weights distributions of WA and BN are smooth and symmetric around zero, while the ones of other normalization methods are rough or asymmetric. Adding WA with LN, IN and GN smooths and symmetrizes the weights distributions of channels. \fig{Chaactivation1} and \fig{Chaactivation2} show activation distributions of a single channel in a convolutional layer. Similarly, the activation distributions of WA and BN are smooth and symmetric around zero, while the ones of other normalization methods are rough or asymmetric. Adding WA with LN, IN and GN smooths and symmetrizes the activation distributions of channels.

\end{document}